\documentclass[runningheads]{llncs}
\usepackage[utf8]{inputenc}

\usepackage{graphicx}

\usepackage[T1]{fontenc}
\usepackage{kantlipsum}
\usepackage[usenames,dvipsnames]{xcolor}
\usepackage[breakable, theorems, skins]{tcolorbox}
\tcbset{enhanced}

\usepackage{url}
\usepackage{graphicx}
\usepackage{hyperref}
\usepackage{booktabs}
\usepackage{caption}
\usepackage{subcaption,blindtext}
\usepackage{multirow}
\usepackage{graphics}
\usepackage{pifont}
\usepackage{color,colortbl}
\usepackage{enumitem}
\usepackage{adjustbox}
\usepackage{placeins}

\usepackage{amsmath,amsfonts,bm}

\def\eqref#1{equation~\ref{#1}}

\def\1{\bm{1}}

\DeclareMathAlphabet{\mathsfit}{\encodingdefault}{\sfdefault}{m}{sl}
\SetMathAlphabet{\mathsfit}{bold}{\encodingdefault}{\sfdefault}{bx}{n}

\usepackage{cleveref}
\usepackage{algorithm}
\usepackage{algcompatible}
\usepackage{amsmath}

\begin{document}
\title{On Web-based Visual Corpus Construction\\for Visual Document Understanding}
\titlerunning{Webvicob for VDU}%

\author{Donghyun Kim\inst{1} \and
Teakgyu Hong\inst{2}\thanks{This work is done at NAVER CLOVA.} \and \\
Moonbin Yim\inst{1} \and
Yoonsik Kim\inst{1} \and
Geewook Kim\inst{1}\thanks{Correspondence to \email{gwkim.rsrch@gmail.com}}}

\authorrunning{Kim et al.}
\institute{NAVER CLOVA \and Upstage AI}
\maketitle              %
\begin{abstract}

In recent years, research on visual document understanding (VDU) has grown significantly, with a particular emphasis on the development of self-supervised learning methods. However, one of the significant challenges faced in this field is the limited availability of publicly accessible visual corpora or extensive collections of images with detailed text annotations, particularly for non-Latin or resource-scarce languages. To address this challenge, we propose Web-based Visual Corpus Builder (Webvicob), a dataset generator engine capable of constructing large-scale, multilingual visual corpora from raw Wikipedia HTML dumps. Our experiments demonstrate that the data generated by Webvicob can be used to train robust VDU models that perform well on various downstream tasks, such as DocVQA and post-OCR parsing. Furthermore, when using a dataset of 1 million images generated by Webvicob, we observed an improvement of over 13\% on the DocVQA Task 3 compared to a dataset of 11 million images from the IIT-CDIP. The implementation of our engine is publicly available on \url{https://github.com/clovaai/webvicob}.

\end{abstract}
\section{Introduction}
Language modeling has been a long-standing fundamental task in natural language processing (NLP). The trained language models (LMs) are utilized in a range of downstream NLP applications, such as information extraction (IE)~\cite{hwang2019pot} and question answering (QA)~\cite{devlinBERT2018}. To build a powerful LM, recent methods utilize large-scale text corpus at the pretraining phase.
The text corpus is generally constructed with a specialized engine or software.
For example, WikiExtractor~\cite{Wikiextractor2015} extracts texts from Wikipedia HTML dumps and builds a clean text corpus.

Visual Document Understanding (VDU)~\cite{xu2019_layoutLM,hong2021bros,kim2022donut} has been developed to conduct a wide range of real-world tasks on document images.
For example, Document Parsing~\cite{xu2019_layoutLM,kim2022donut} aims to extract some key texts from a document image~\cite{hwang2019pot,hwang2021costeffective}. Most recent VDU backbones can be regarded as an extension of LM.

\begin{figure}[htp]
    \centering
    \begin{minipage}{\linewidth}
        \centering
        \begin{subfigure}{0.24\textwidth}
            \captionsetup{oneside,margin={0.0cm,0cm},font=tiny}
            \includegraphics[width=\textwidth]{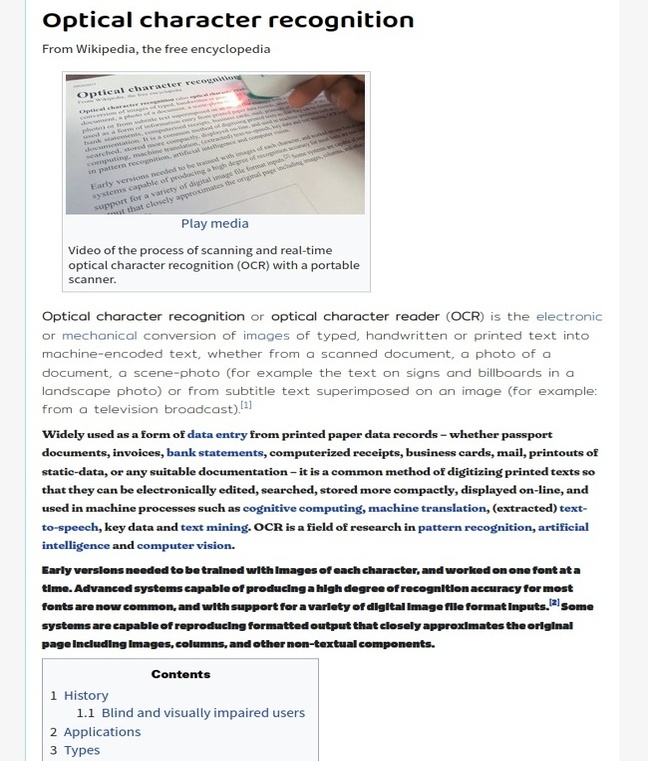}
            \subcaption*{Webvicob}
        \end{subfigure}
        \begin{subfigure}{0.24\textwidth}
            \captionsetup{oneside,margin={0.0cm,0cm},font=tiny}
            \includegraphics[width=\textwidth]{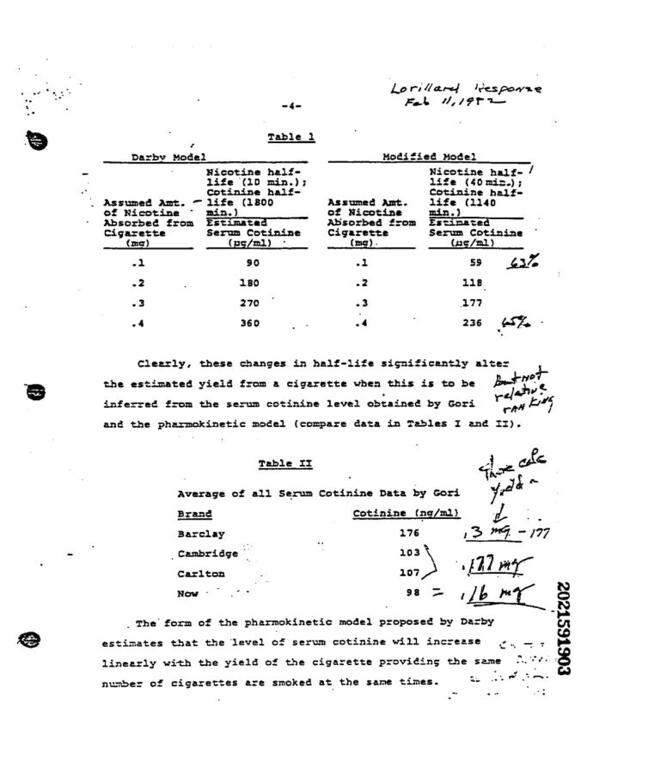}
            \subcaption*{IIT-CDIP}
        \end{subfigure}
        \begin{subfigure}{0.24\textwidth}
            \captionsetup{oneside,margin={0.0cm,0cm},font=tiny}
            \includegraphics[width=\textwidth]{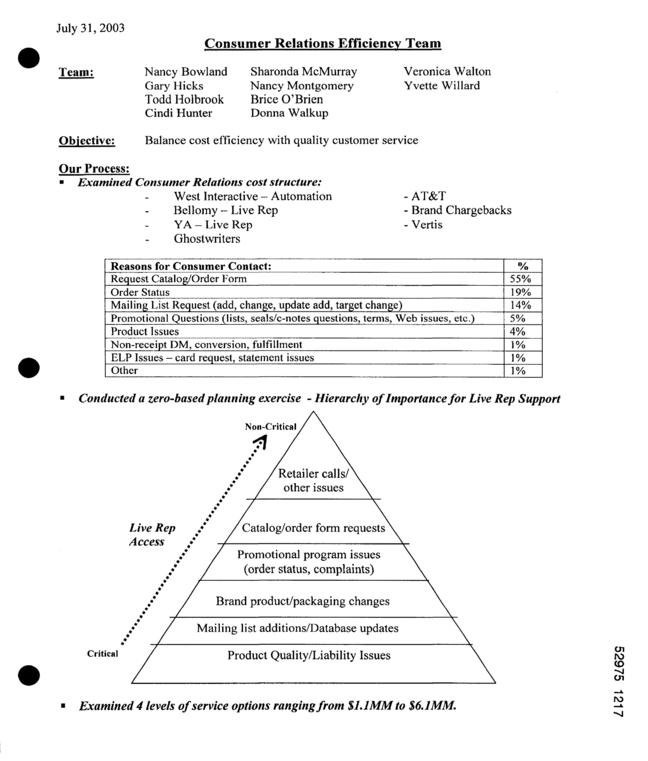}
            \subcaption*{DocVQA Task 1}
        \end{subfigure}
        \begin{subfigure}{0.24\textwidth}
            \captionsetup{oneside,margin={0.0cm,0cm},font=tiny}
            \includegraphics[width=\textwidth]{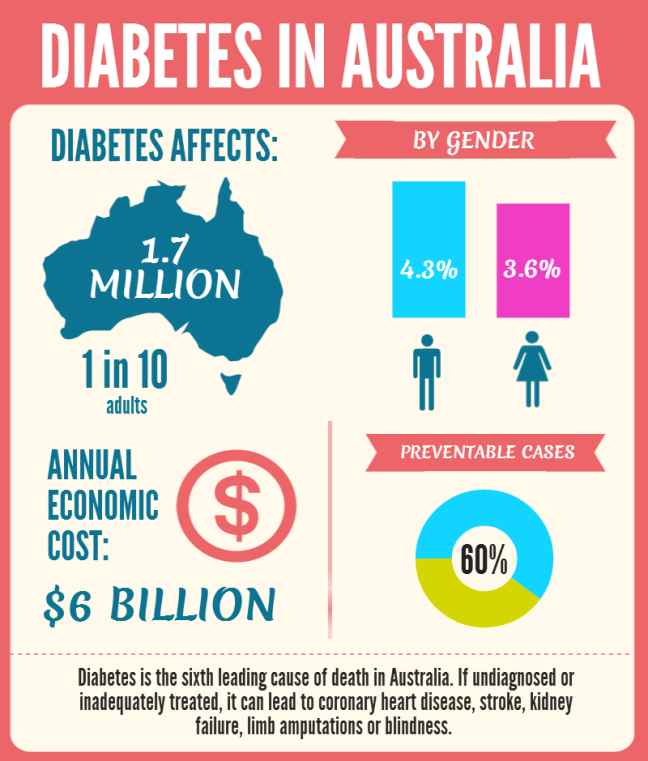}
            \subcaption*{DocVQA Task 3}
        \end{subfigure}
        \hfill

        \subcaption{Document image samples.}
        \label{fig:1a}
    \end{minipage}
    \begin{minipage}{\linewidth}
        \centering
        \begin{subfigure}{0.24\textwidth}
            \captionsetup{oneside,margin={0.0cm,0cm},font=tiny}
            \includegraphics[width=\textwidth]{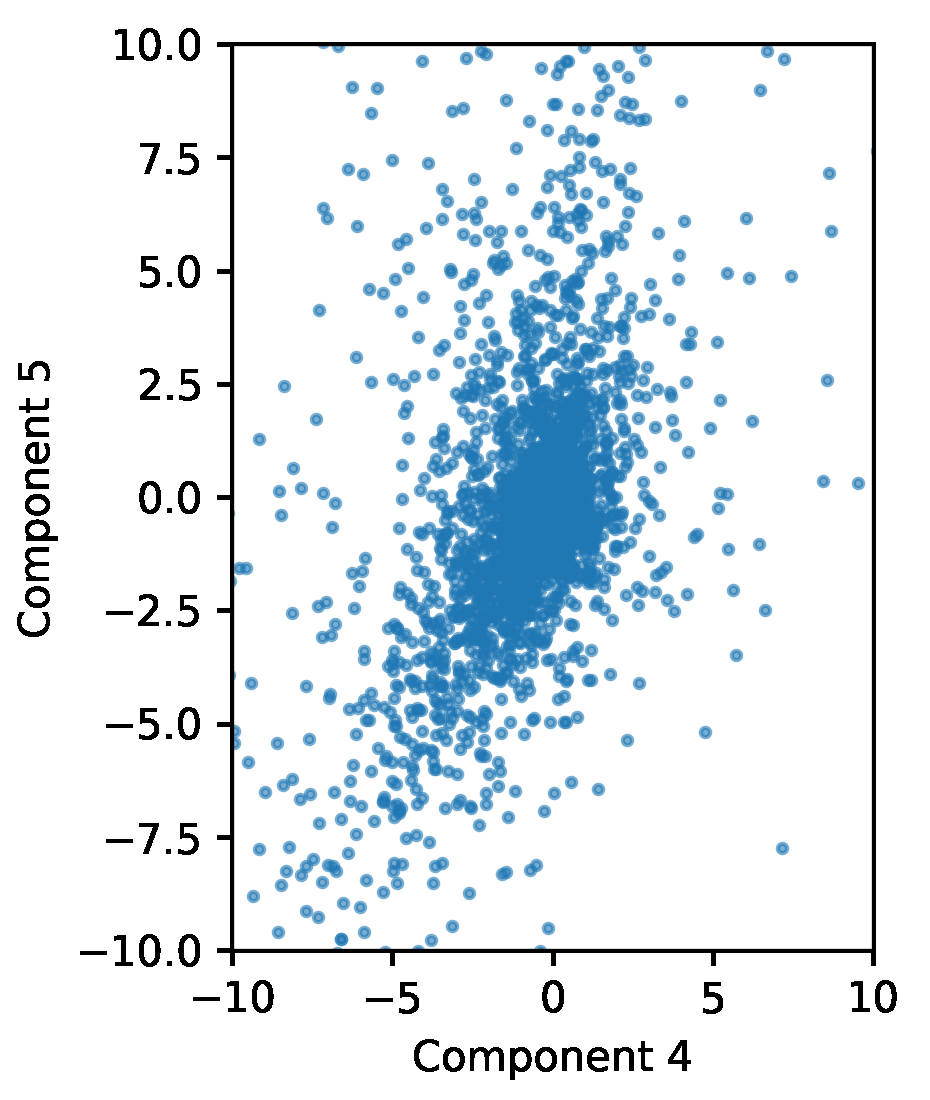}
            \subcaption*{Webvicob}
        \end{subfigure}
        \begin{subfigure}{0.24\textwidth}
            \captionsetup{oneside,margin={0.0cm,0cm},font=tiny}
            \includegraphics[width=\textwidth]{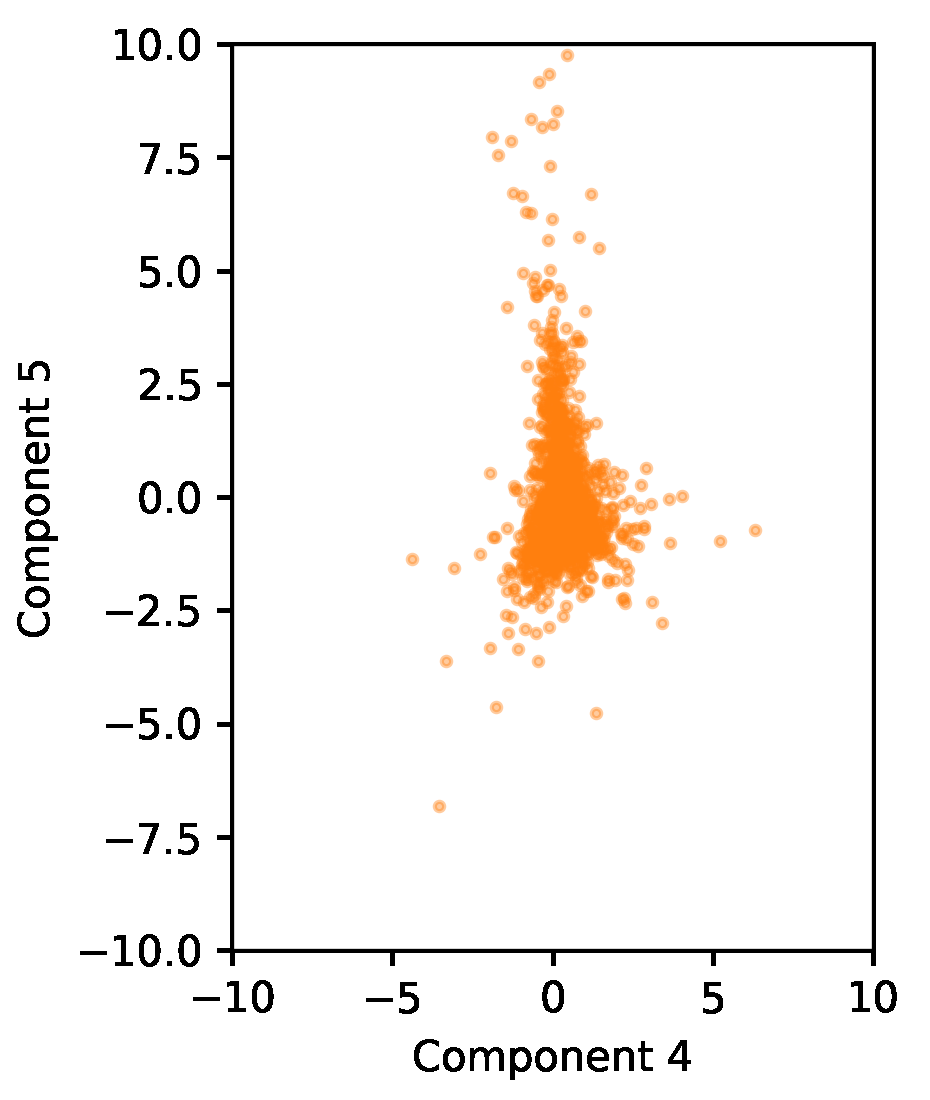}
            \subcaption*{IIT-CDIP}
        \end{subfigure}
        \begin{subfigure}{0.24\textwidth}
            \captionsetup{oneside,margin={0.0cm,0cm},font=tiny}
            \includegraphics[width=\textwidth]{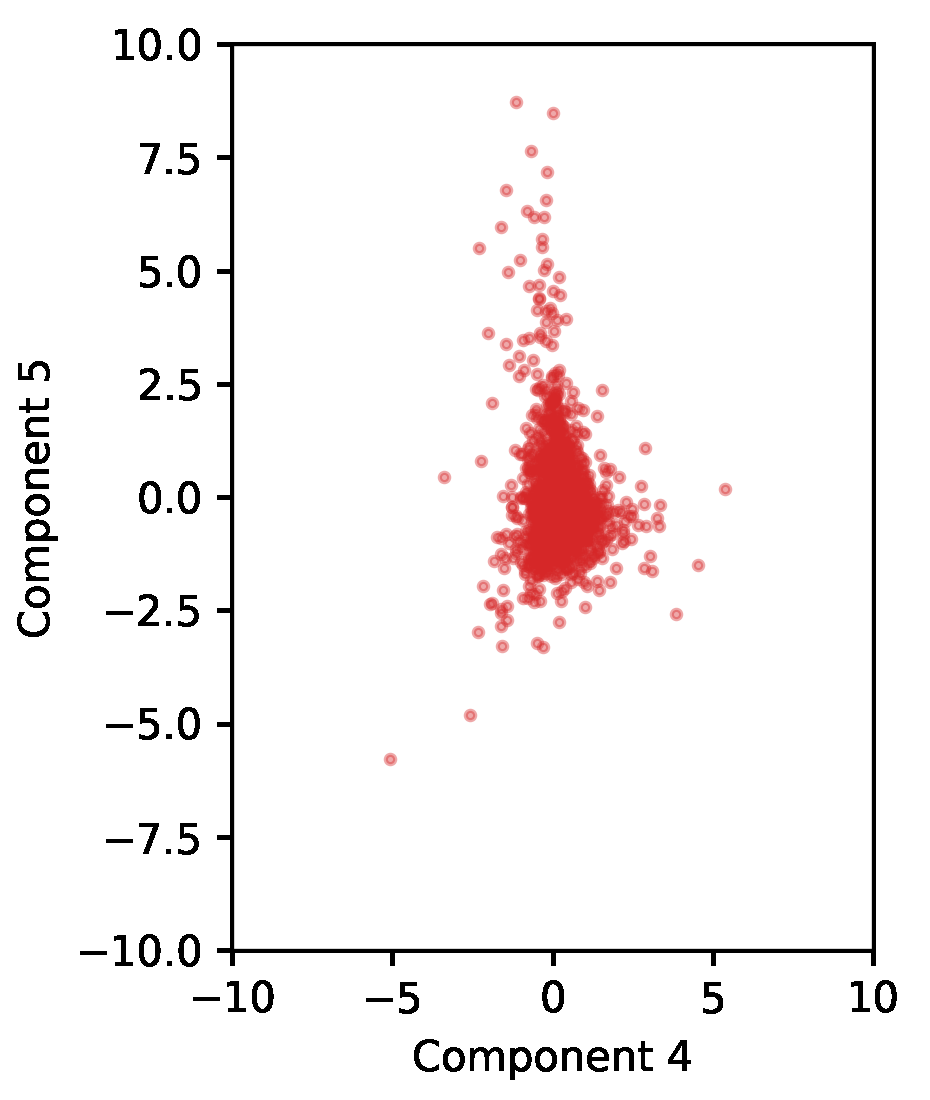}
            \subcaption*{DocVQA Task 1}
        \end{subfigure}
        \begin{subfigure}{0.24\textwidth}
            \captionsetup{oneside,margin={0.0cm,0cm},font=tiny}
            \includegraphics[width=\textwidth]{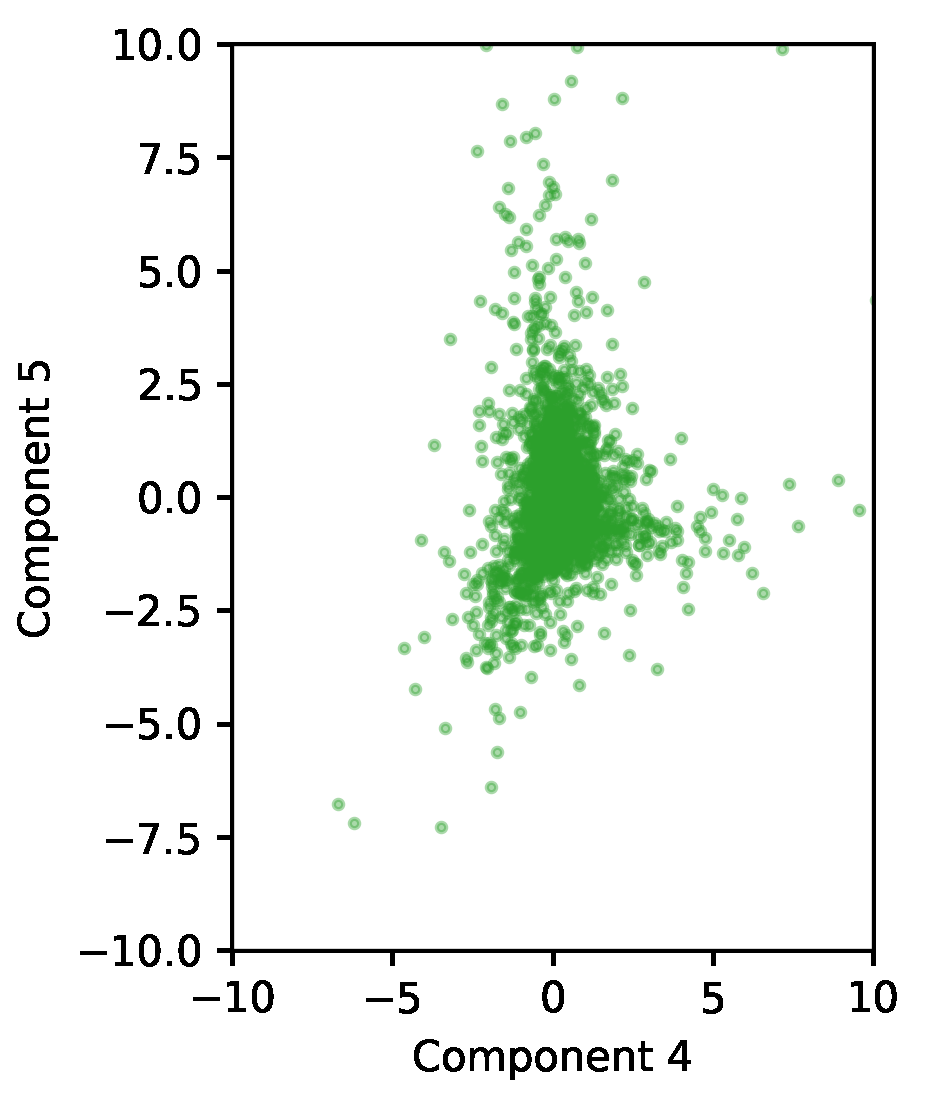}
            \subcaption*{DocVQA Task 3}
        \end{subfigure}
        \hfill

        \subcaption{PCA visualization.}
        \label{fig:1b}
    \end{minipage}
    \begin{minipage}{\linewidth}
        \centering
        \includegraphics[width=0.49\linewidth]{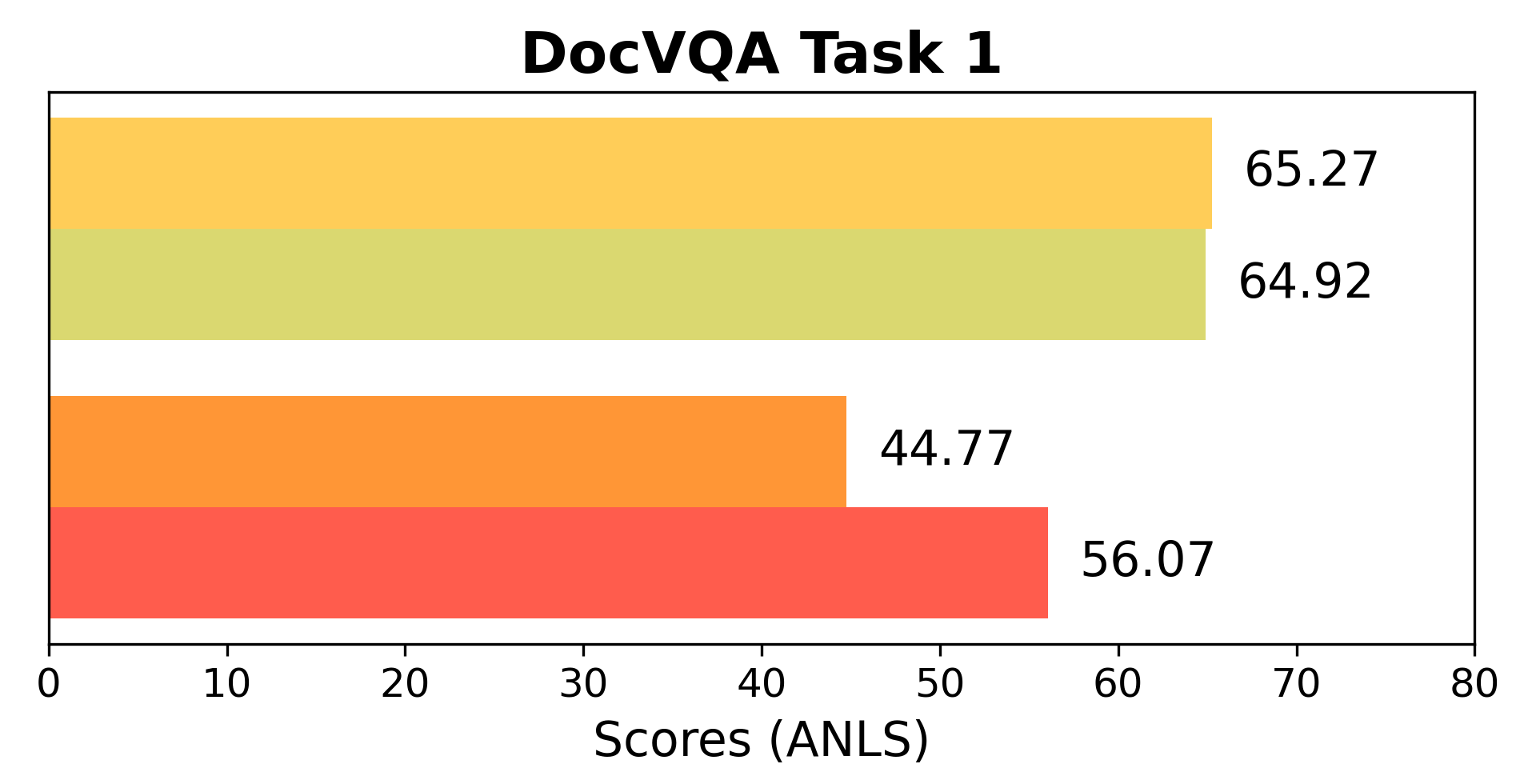}
        \includegraphics[width=0.49\linewidth]{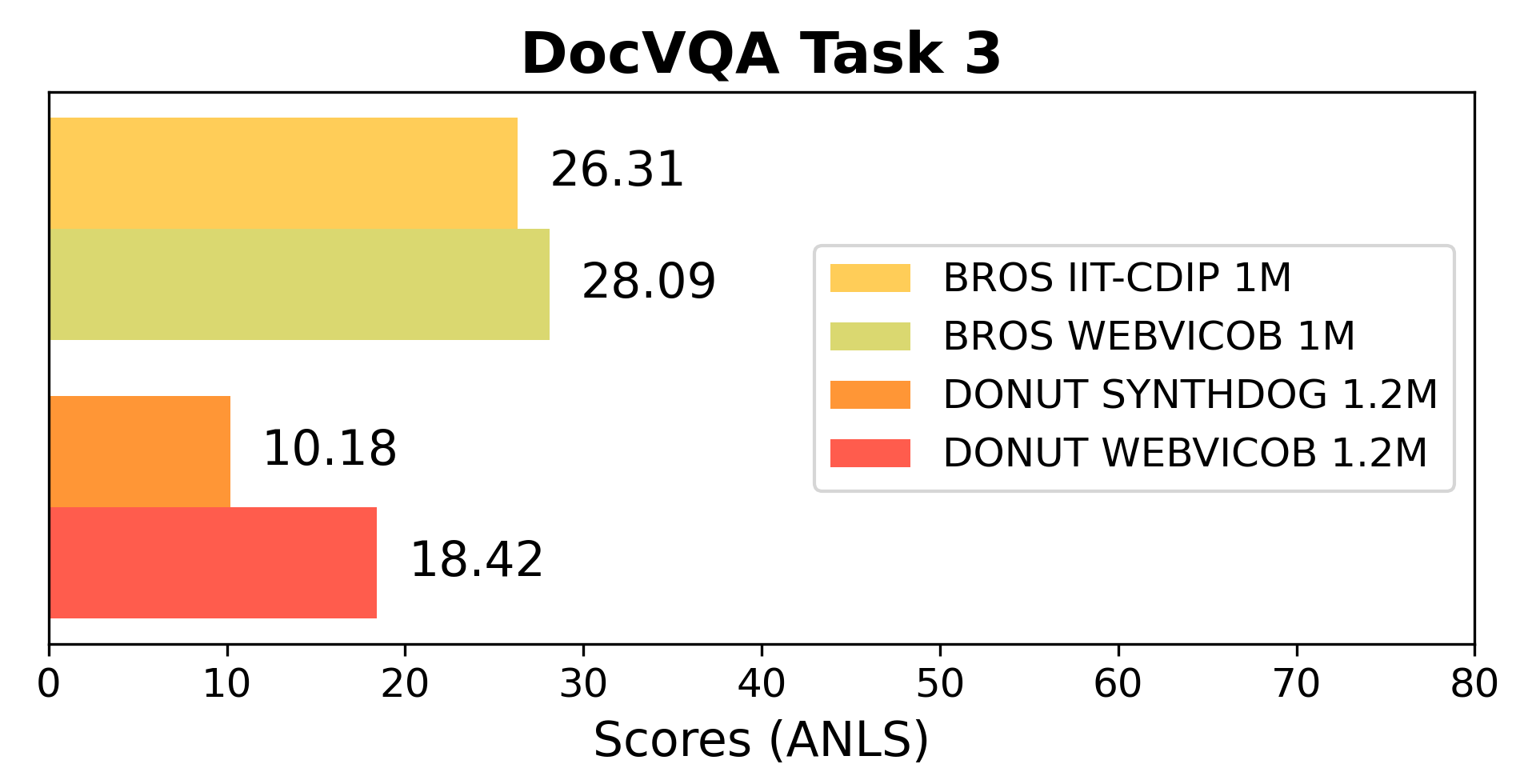}
        \subcaption{Document VQA benchmarks.}
        \label{fig:1c}
    \end{minipage}
    \caption{\textbf{Image Samples and Visualization of Main Results.} (a) Samples of Webvicob, IIT-CDIP~\protect\cite{iitcdip}, DocVQA Task 1~\protect\cite{icdar21docvqa}, and Task 3~\protect\cite{mathew2022infographicvqa} are shown, respectively. (b) Visualization of principal component analysis (PCA) with bag of words (BoW) representations. The visualization was carried out using the $4^{th}$ and $5^{th}$ principal components. The analysis method and further visualization results can be found in section~\ref{sec:pca_analysis}. (c) Results on two benchmarks with pretrained BROS$_{\text{BASE}}$~\protect\cite{hong2021bros} and Donut$_{\text{Proto}}$~\protect\cite{kim2022donut} are shown. As can be seen in (a) and (b), DocVQA Task 1 has a similar distribution to IIT-CDIP, while Task 3 does not. This also affects the final score.}
    \label{fig:1}
\end{figure}

\begin{figure*}[t]
    \centering
    \includegraphics[width=0.95\textwidth]{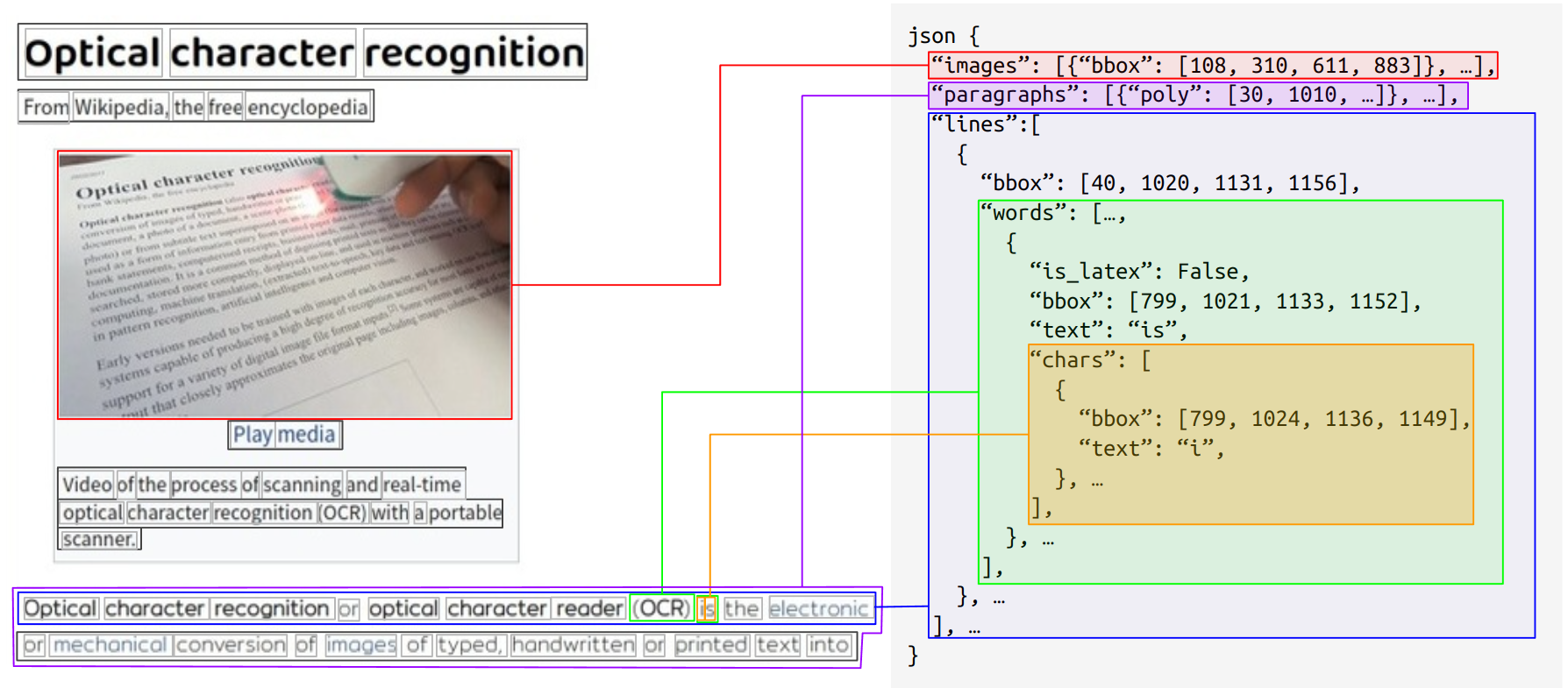}
    \caption{\textbf{A sample of Webvicob dataset.} The dataset contains large-scale web document images with hierarchical text annotations (i.e., character, word, line, and paragraph-level annotations). If box contains LaTeX~\cite{latex} (i.e., math formula), we set ``{is\_latex}'' value as True.}
    \label{fig:2}
\end{figure*}

\begin{figure}[t]
    \centering
    \begin{subfigure}{0.24\textwidth}
        \includegraphics[width=\textwidth]{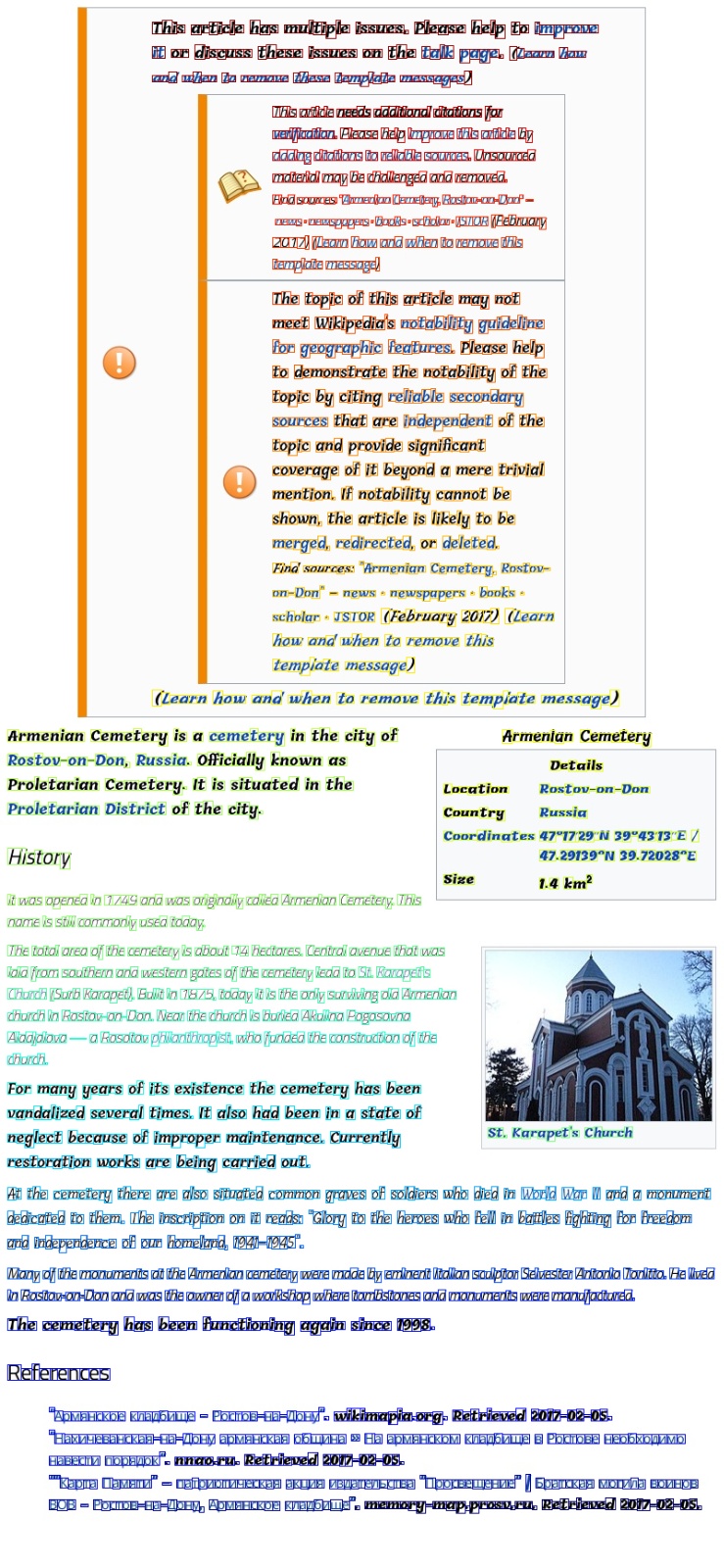}
        \caption{Character}
        \label{fig:4(a)}
    \end{subfigure}
    \begin{subfigure}{0.24\textwidth}
        \includegraphics[width=\textwidth]{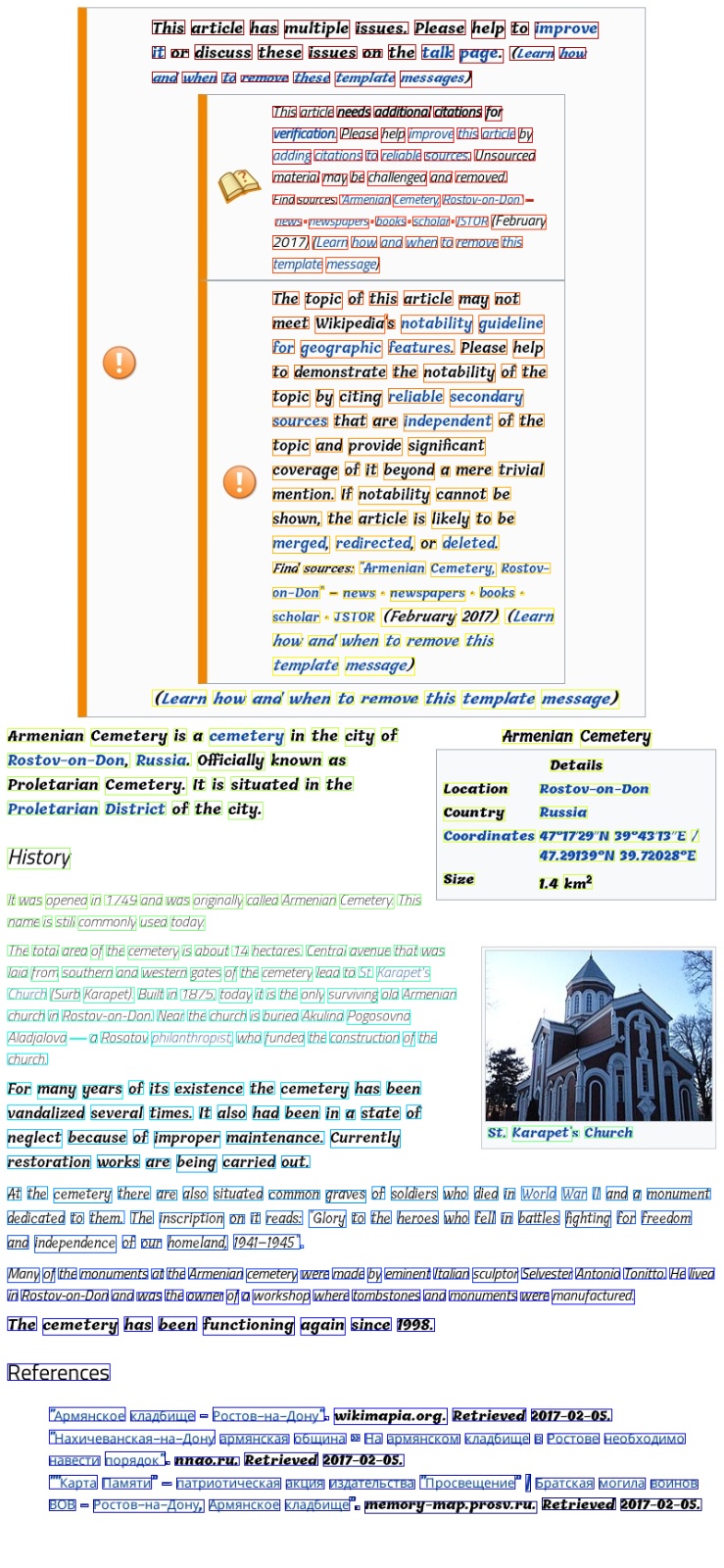}
        \caption{Word}
        \label{fig:4(b)}
    \end{subfigure}
    \begin{subfigure}{0.24\textwidth}
        \includegraphics[width=\textwidth]{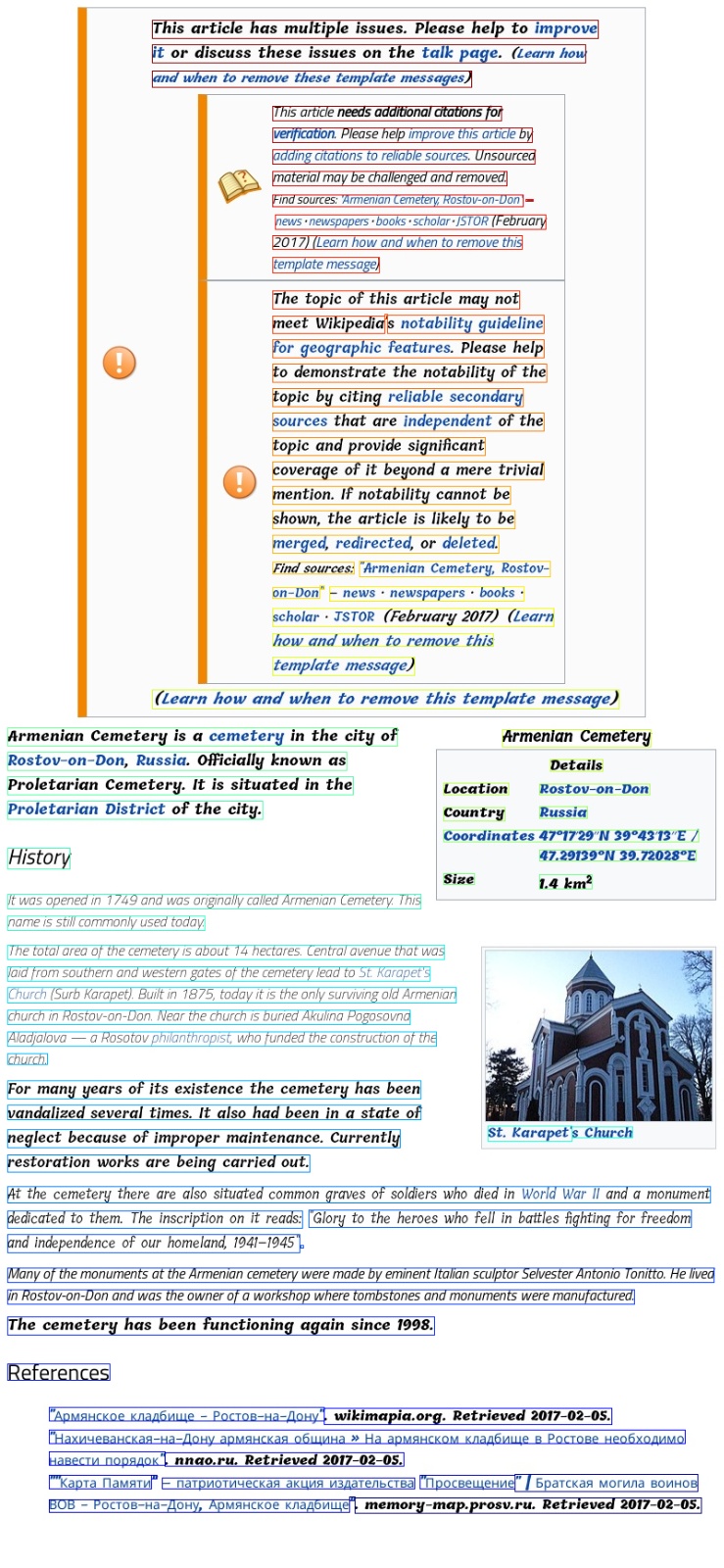}
        \caption{Line}
        \label{fig:4(c)}
    \end{subfigure}
    \begin{subfigure}{0.24\textwidth}
        \includegraphics[width=\textwidth]{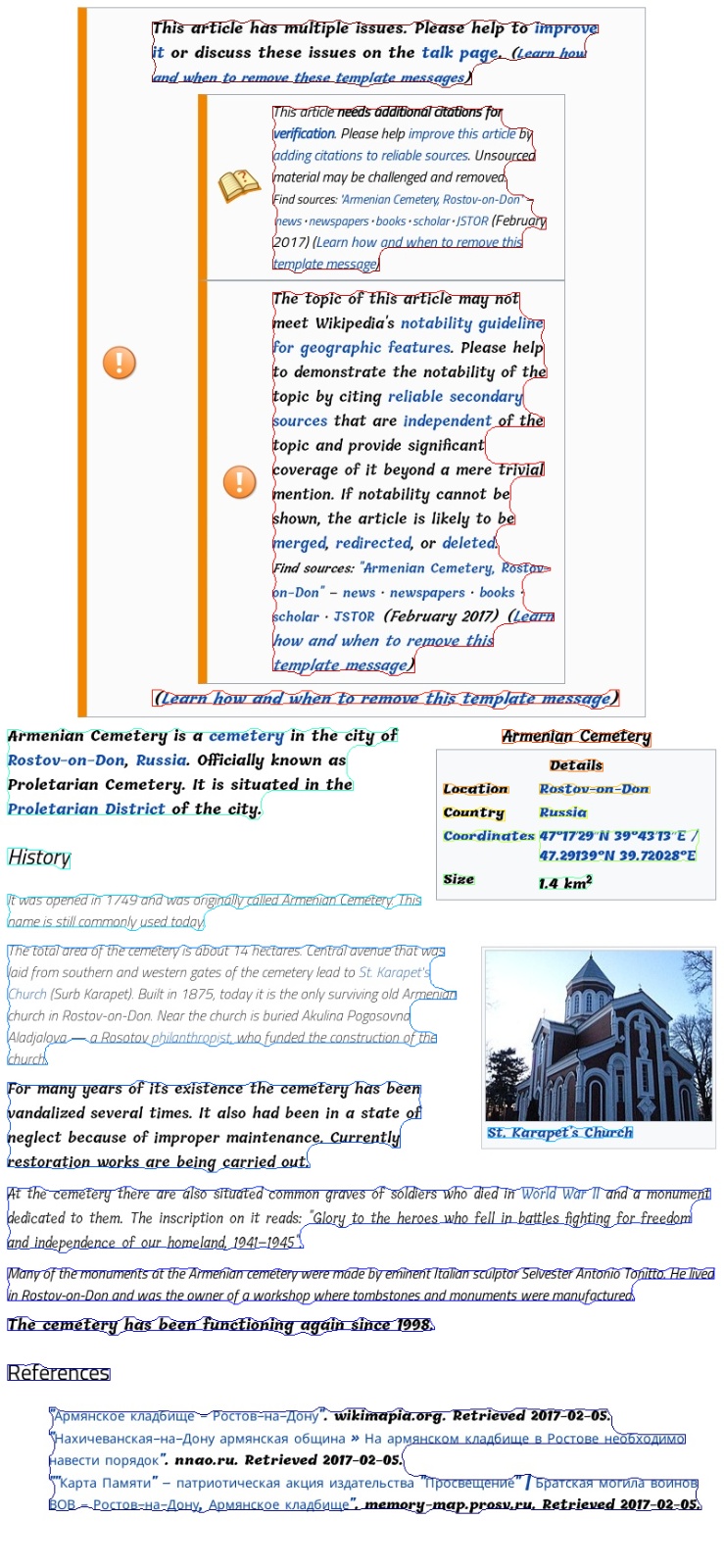}
        \caption{Paragraph}
        \label{fig:4(d)}
    \end{subfigure}

    \caption{\textbf{Visualization of Webvicob annotations.} We use a colormap to show the order of box annotation.}
    \label{fig:viz_annot1}
\end{figure}

Inspired by recent advances in LMs~\cite{vaswani2017transformer,devlinBERT2018},
recent VDU methods share a similar approach that (1) it first collects large-scale real document images, (2) conducts OCR on the images to extract texts, and (3) trains a BERT-like LM backbone on the extracted texts~\cite{xu2019_layoutLM,xu2021layoutlmv2,hong2021bros,huang2022layoutlmv3}.

Although conventional VDU methods have shown promising results, several practical challenges exist, especially in the training dataset preparation phase.
Most recent works~\cite{xu2019_layoutLM,hong2021bros} rely on a large-scale real image dataset. For example, IIT-CDIP dataset~\cite{iitcdip} consists of 11M industrial document images and is utilized in a range of VDU works.
However, in most low-resourced languages, there is no public dataset like IIT-CDIP. 
In addition, off-the-shelf OCR engines (e.g., CLOVA OCR API,\footnote{\scriptsize\label{clova_OCR_note}\url{https://clova.ai/OCR}} MS Read API,\footnote{\scriptsize\url{https://docs.microsoft.com/en-us/azure/cognitive-services/computer-vision}} Amazon Textract\footnote{\scriptsize\url{https://learn.microsoft.com/ko-kr/azure/cognitive-services/computer-vision/overview-ocr}}) are required to extract texts in the pre-processing.
This often requires enormous costs.

Moreover, using OCR can have other negative consequences; the constructed data is strongly tied to the OCR engine, and OCR errors are propagated throughout the process. This problem becomes severe, especially in some non-Latin languages such as Korean and Japanese, where OCR is known to be complicated.

In this work, we propose an engine for building a web-based visual corpus. As shown in Figure~\ref{fig:2} and Figure~\ref{fig:viz_annot1}, the proposed Web-based Visual Corpus Builder (Webvicob) renders a web page into an image file and generates rich text annotations. With expressive Document Object Model (DOM) APIs, Webvicob produces accurate bounding boxes for all characters.

Moreover, Webvicob covers a wide range of word contexts. Wikipedia is a huge dataset consisting of over 270 languages and containing 60 million HTML documents. The results of PCA analysis on 3,626 samples are available in Section~\ref{sec:pca_analysis}.

Compared to the traditional model trained on IIT-CDIP, the Webvicob-trained model shows a competitive result on DocVQA Task 1 and a higher score\footnote{\scriptsize\url{https://rrc.cvc.uab.es/?ch=17&com=evaluation&task=3}} on Task 3, showing the effectiveness of the de-biased Webvicob-based visual corpora.

The proposed method is simple yet effective. We show that the Webvicob-generated corpus is critical in building a powerful VDU backbone.
Through extensive experiments and analyses, we show the effectiveness of Webvicob.
The contributions of this work can be summarized as follows:
\begin{enumerate}
    \item We propose Webvicob, which can be used for pretraining the visual document understanding models. Webvicob provides rich annotations, including character, word, line, and paragraph information. 
    \item Webvicob provides support for a wide range of fonts, allowing visual documents with identical content to appear differently. Additionally, we have taken into account the characteristics of each font to construct precise character-level bounding box annotations.
    \item We conduct extensive experiments to verify the effectiveness of the proposed engine and dataset.
    \item The source code of our engine is publicly available to promote research on VDUs in low-resource languages.
\end{enumerate}

\section{Background and Related Work}
In this section, we introduce a traditional VDU pipeline and datasets. Most of the current methods share a similar approach of (1) collecting large-scale real images, (2) conducting OCR on the images, and (3) training a BERT-like backbone on the extracted texts~\cite{xu2019_layoutLM,hong2021bros}.

\subsection{VDU Backbones}
Inspired by BERT~\cite{devlinBERT2018} and the recent advancements in language modeling, a range of BERT-like Transformer-based VDU backbones have been proposed \cite{xu2019_layoutLM,hong2021bros,xu2021layoutlmv2,huang2022layoutlmv3,DOM_LM}.
For handling layout information of document images, 
spatial coordinates of OCR text boxes are fed to the VDU backbone~\cite{xu2019_layoutLM,hong2021bros,layoutxlm}.
Using visual encoders like ResNet~\cite{HeZRS16}, visual features of an input image are also being incorporated into the recent VDU backbones~\cite{xu2021layoutlmv2,huang2022layoutlmv3}. More recently, with the advances in Vision Transformer (ViT)~\cite{dosovitskiy2020vit}, training a Transformer encoder-decoder VDU backbone without OCR has also been attempted~\cite{kim2022donut,Dessurt,pix2struct}.
Our engine can be used together in various VDU backbones. In this paper, we verified the performance of the Webvicob engine by pretraining BROS~\cite{hong2021bros}, LayoutXLM~\cite{layoutxlm}, and Donut~\cite{kim2022donut}.

\subsection{Visual Corpus Construction for VDU}
Most existing OCR annotated datasets have small sizes, leading to difficulties in training VDU backbones.
To construct a rich corpus, in the traditional pipeline, large-scale real-world document images (e.g., IIT-CDIP) and an OCR engine (e.g., CLOVA OCR API~\textsuperscript{\ref{clova_OCR_note}}) are used.

The quality of the OCR engine significantly affects the downstream processes~\cite{kim2022donut,Dessurt}.
Hence, there have been difficulties in training and testing the VDU backbone. For example, since BROS~\cite{hong2021bros} and LayoutLM~\cite{layoutxlm} use different in-house OCRs, it has been difficult to make a fair comparison.

LayoutXLM~\cite{layoutxlm} collects large-scale digital-born PDF data from the world wide web and extracts text annotations from the PDF via an open-source PDF renderer.\footnote{\scriptsize\label{pymupdf}\url{https://github.com/pymupdf/PyMuPDF}}
Although this showed another promising direction, it is not easy to follow the pipeline in practice. A practitioner has to collect the PDF data as there is no publicly available dataset (\cite{layoutxlm} did not open-source the dataset).

Moreover, since PDF files cannot easily be editable, augmentation is limited.
Unlike the existing approach, Webvicob can efficiently modify and augment data (i.e., layout, background image) with javascript as Webvicob renders HTML directly.
Moreover, Webvicob can easily be incorporated with the HTML dumps (e.g., Wikipedia dumps\footnote{\scriptsize\url{https://dumps.wikimedia.org}}), which are easily accessible and have been widely used in building powerful NLP backbones~\cite{devlinBERT2018}.

\section{Web-based Visual Corpus Builder}
Webvicob uses HTML dumps and modifies Document Object Model (DOM) to generate data for pretraining VDU backbones with rich corpora.
As seen in Figure~\ref{fig:2}, Webvicob provides box annotations of characters, words, LaTeX~\cite{latex}, images, lines, and paragraphs, and also for images and LaTeXs.

In this section, we explain the generation procedures (Algorithm~\ref{alg:ProcessHTML}) in detail.

\renewcommand{\algorithmicrequire}{\textbf{Input:}}
\renewcommand{\algorithmicensure}{\textbf{Output:}}
\algnewcommand\algorithmicreturn{\textbf{return}}
\algnewcommand\RETURN{\algorithmicreturn}
\algnewcommand\algorithmicprocedure{\textbf{procedure}}
\algnewcommand\PROCEDURE{\item[\algorithmicprocedure]}%
\algnewcommand\algorithmicendprocedure{\textbf{end procedure}}
\algnewcommand\ENDPROCEDURE{\item[\algorithmicendprocedure]}%
\algnewcommand{\algvar}[1]{{\text{\ttfamily\detokenize{#1}}}}
\algnewcommand{\algstart}[1]{{\text{\ttfamily\detokenize{#1}}}}
\algnewcommand{\algarg}[1]{{\text{\ttfamily\itshape\detokenize{#1}}}}
\algnewcommand{\algproc}[1]{{\text{\color{blue}\ttfamily\detokenize{#1}}}}
\algnewcommand{\algassign}{\leftarrow}

\begin{algorithm}[h!]
\footnotesize
\caption{Get annotations from HTML}
\label{alg:ProcessHTML}
\begin{algorithmic}[1]
\REQUIRE \algvar{html}
\ENSURE \algvar{image, annotations}
\Statex
\PROCEDURE \algstart{get_annotations_from_html}(\algvar{html})
  \STATE $\algvar{html} \algassign \algproc{add_spans}(\algvar{html})$ \Comment{\ref{alg:add_span}}
  \Statex // From $\scriptstyle \textbf{\textless p\textgreater ab\textless/p\textgreater }$ to $\scriptstyle \textbf{\textless p\textgreater \textless span\textgreater a\textless/span\textgreater \textless span\textgreater b\textless/span\textgreater \textless/p\textgreater }$
  \STATE $\algvar{driver} \algassign \algproc{get_selenium_driver}(\algvar{html})$
  \STATE $\algproc{remove_unusable_elements}(\algvar{driver})$ \Comment{\ref{alg:preprocess}}
  \STATE $\algproc{change_paragraph_fonts}(\algvar{driver})$ \Comment{\ref{alg:para}}
  \STATE $\algvar{image} \algassign \algproc{capture}(\algvar{driver})$
  \STATE $\algvar{boxes} \algassign \algproc{get_glyph_box}(\algvar{driver})$ \Comment{\ref{alg:bbox}}
  \STATE $\algvar{annotations} \algassign \algproc{get_annots}(\algvar{boxes})$ \Comment{\ref{alg:annot}}
  
  \Statex \RETURN{} \algvar{image, annotations}
\end{algorithmic}
\end{algorithm}

\subsection{Annotation}
\subsubsection{Adding Spans}
\label{alg:add_span}
We can access each Element\footnote{\scriptsize\url{https://developer.mozilla.org/en-US/docs/Web/API/Element}} using the DOM and find out the bounding box (bbox) of the Element through the \texttt{getBoundingClientRect()}\footnote{\scriptsize\url{https://developer.mozilla.org/en-US/docs/Web/API/Element/getBoundingClientRect}} function. Webvicob modifies the HTML so that all characters can be bounded with a \textless span\textgreater  tag to get the bbox of each character (i.e., from {\scriptsize \textbf{\textless p\textgreater abc\textless/p\textgreater }} to {\scriptsize \textbf{\textless p\textgreater \textless span\textgreater a\textless/span\textgreater \textless span\textgreater b\textless/span\textgreater \textless span\textgreater c\textless/span\textgreater \textless/p\textgreater }}). With this procedure, the \texttt{getBoundingClientRect()} function can be applied to all spans, allowing us to obtain bounding box annotations for all characters.

\subsubsection{Remove Unusable Elements} 
\label{alg:preprocess}
The essential step before creating data is to remove unusable Elements. For example, there are various ways to hide a specific Element in a web document. Even if the Element is invisible, the function \texttt{getBoundingClientRect()} returns results since Element occupies space. Specifically, the following Elements are removed:
\let\labelitemi\labelitemii
\begin{itemize}
    \item Pseudo Elements.\footnote{\scriptsize{\url{https://developer.mozilla.org/en-US/docs/Web/CSS/Pseudo-elements}}}
    \item Child Elements whose Element size is larger than the parent node.
    \item Elements that invisible style applied. \\
          (display: none / visibility: hidden / visibility: collapse / opacity: 0)
    \item Elements located outside the rendering screen.
    \item Element with placeholder attribute.
\end{itemize}

\subsubsection{Construct Annotations} 
\label{alg:annot}
We construct word annotations and line annotations by calculating the spaces between the character boxes and the line boxes. We define LaTex Elements with ``mwe-math-fallback-image-inline'' className, and define image Elements with \{image, canvas, SVG, video\} tags. As MarkupLM~\cite{MarkupLM} did, paragraph annotations are extracted using a well-ordered tree structure. Elements with the same depth are grouped into one paragraph.

\subsection{Rendering with Various Fonts}

\subsubsection{Random Paragraph Fonts} 
\label{alg:para}

Visual diversity in pretraining datasets is generally associated with improved performance of VDU backbones~\cite{mjsynth,synthtext,synthtiger,kim2022donut}.

\begin{figure}[h]
    \centering
    \begin{subfigure}{0.3\columnwidth}
        \includegraphics[width=\columnwidth]{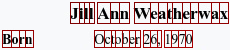}
        \caption{}
        \label{fig:3(a)}
    \end{subfigure}
    \hfill
    \begin{subfigure}{0.3\columnwidth}
        \includegraphics[width=\columnwidth]{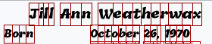}
        \caption{}
        \label{fig:3(b)}
    \end{subfigure}
    \hfill
    \begin{subfigure}{0.3\columnwidth}
        \includegraphics[width=\columnwidth]{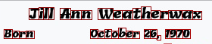}
        \caption{}
        \label{fig:3(c)}
    \end{subfigure}
    
    \caption{\textbf{Visualization of rendered images with various fonts.} (a) Original font with \texttt{getBoundingClientRect()} results. (b) Random font with \texttt{getBoundingClientRect()} results. (c) Random font with actual glyph boxes.}
\end{figure}

For visual diversity, Webvicob renders HTML with various fonts for each paragraph (See Figure~\ref{fig:3(a)} and \ref{fig:3(b)}). We randomly select fonts from 3,567 GoogleFonts\footnote{\scriptsize\url{https://fonts.google.com}} in our experiments and analyses.

\subsubsection{Precise Bounding Boxes}
\label{alg:bbox}

The actual bounding box of the glyph and the result of \texttt{getBoundingClientRect()} are different. As shown in Figure~\ref{fig:3(a)} and \ref{fig:3(b)}, the result of \texttt{getBoundingClientRect()} has a large margin. We extract a ratio of the actual glyph box to the bounding box via rendering vector images in a font file with a Pygame FreeType handler.\footnote{\scriptsize\url{https://www.pygame.org/docs/ref/freetype.html}} Using the ratio, the final tight glyph box can be obtained (Figure~\ref{fig:3(c)}).

\section{Experiment and Analysis}
\subsection{Setup}
Donut experiments are carried out using 8 NVIDIA V100 GPUs for a fair comparison with Donut$_{\text{Proto}}$, while other experiments are conducted using 4 NVIDIA A100 80G GPUs. We use mixed precision training technique~\cite{MixedPrecision}. 

\subsubsection{Employed Models}
To show the effectiveness of Webvicob, We use three models with different properties.

BROS~\cite{hong2021bros} was proposed with an effective pretraining method (i.e., area masking) and a relative positional encoding. To validate the effectiveness of Webvicob-generated data, we pretrain BROS$_{\text{BASE}}$ and measure performance on DocVQA Task 1~\cite{icdar21docvqa} and Task 3~\cite{mathew2022infographicvqa}.

LayoutXLM~\cite{layoutxlm} is a widely-used multilingual VDU backbone. We re-implement and pretrain LayoutXLM$_{\text{BASE}}$ for eight languages to validate Webvicob in a multilingual scenario. FUNSD~\cite{jaume2019funsd} and XFUND~\cite{layoutxlm} datasets are used as benchmark datasets.

Donut~\cite{kim2022donut} introduced a novel approach that utilizes images alone without relying on OCR results as input. We pretrain Donut$_{\text{Proto}}$ using data generated by Webvicob to demonstrate the versatility of the data for different architectures. This has been verified through experiments on DocVQA Task 1, DocVQA Task 3, FUNSD, and XFUND (Japanese) datasets. 

\subsubsection{Donut for Entity Recognition}
\begin{figure*}[h!]
    \centering
    \includegraphics[width=0.95\textwidth]{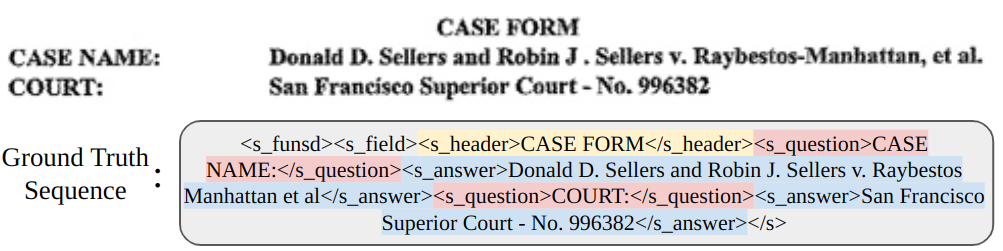}
    \caption{\textbf{A sample of ground truth for Donut Entity Recognition.}}
    \label{fig:task_explain}
\end{figure*}

Since Donut only takes an image input and cannot utilize the popular method of token classification using OCR information, we trained the model to decode sequences in a specific format (as depicted in Figure~\ref{fig:task_explain}) in order to extract all entity fields within the document.

To assess the model's ability to identify the entity set, regardless of the order, we used Hungarian Matching~\cite{Kuhn1955Hungarian} to adjust the sequence so that the Tree Edit Distance (TED) is minimized. The final evaluation score was based on TED-based accuracy~\cite{TED1,hwang2021costeffective,TED3,kim2022donut}.

\begin{table*}[t]
\caption{\textbf{Statistics of Webvicob and conventional datasets.} ISO 639-1 language code is used for Language Support. Webvicob$^\dagger$ is used in our multilingual experiments (Section~\ref{sec:multi}). Webvicob$^\ddagger$ is a virtual dataset that can be constructed with Webvicob. It contains more than 60M samples with rich annotations and 270+ language supports. The total number of Wikipedia articles (60M) can be found on the Wikipedia statistic webpage.\protect\footnotemark}
\centering
\resizebox{1.0\textwidth}{!}{%
    \begin{tabular}{crrrrrrrrl}
    \toprule
    \multirow{2}{*}{Dataset} & \multicolumn{3}{c}{\#Image} & \multicolumn{1}{c}{} & \multicolumn{4}{c}{Annotation Level} & \multirow{2}{*}{Language Supports} \\
    \cline{2-9}
     & \multicolumn{1}{c}{Train} & \multicolumn{1}{c}{Val} & \multicolumn{1}{c}{Test} & \multicolumn{1}{c}{} & \multicolumn{1}{c}{Char} & \multicolumn{1}{c}{Word} & \multicolumn{1}{c}{Line} & \multicolumn{1}{c}{Para} &  \\
     \toprule
    MSRA-TD500~\cite{msratd500} & 300 & 0 & 200 & &  &  & \ding{51} &  & EN \\
    IC15~\cite{ic15} & 1,000 & 0 & 500 & &  & \ding{51} &  &  & EN \\
    CTW-1500~\cite{ctw1500} & 1,000 & 0 & 500 & &  &  & \ding{51} &  & EN \\
    IC17 MLT~\cite{ic17mlt} & 7,200 & 1,800 & 9,000 & &  & \ding{51} &  &  & EN ZH JA KO BN AR \\

    IC19 MLT~\cite{ic19mlt} & 10,000 & 0 & 10,000 & &  & \ding{51} &  &  & EN ZH JA KO BN AR HI \\

    IC19 LSVT~\cite{ic19lsvt} & 30,000 & 0 & 20,000 & &  &  & \ding{51} &  & EN ZH \\
    IC19 ArT~\cite{ic19art} & 5,603 & 0 & 4,563 & &  & \ding{51} &  &  & EN ZH \\
    Total-Text~\cite{totaltext} & 1,255 & 0 & 300 & &  & \ding{51} &  &  & EN ZH \\
    TextOCR~\cite{textOCR} & 21,778 & 3,124 & 3,232 & &  & \ding{51} &  &  & EN \\
    IntelOCR~\cite{intelOCR} & 191,059 & 16,731 & 0 & &  & \ding{51} &  &  & EN \\
    HierText~\cite{hiertext} & 8,281 & 1,724 & 1,634 & &  & \ding{51} & \ding{51} & \ding{51} & EN \\ \hline
    SynthText~\cite{synthtext} & 858,750 & 0 & 0 & & \ding{51} & \ding{51} &  &  & EN \\ \hline
    IIT-CDIP~\cite{iitcdip} & 11,434,146 & 0 & 0 & &  &  &  &  & EN \\
    OCR-IDL~\cite{OCR_IDL} & 26,600,964 & 0 & 0 & &  & \ding{51} & \ding{51} &  & EN \\ \hline
    Webvicob$^\dagger$ & 18,584,173 & 4000 & 4000 & & \ding{51} & \ding{51} & \ding{51} & \ding{51} & EN ZH JA ES FR PT IT DE \\

    Webvicob$^\ddagger$ & \multicolumn{3}{c}{60,475,636} & & \ding{51} & \ding{51} & \ding{51} & \ding{51} & 270+ languages \\ \bottomrule
    \end{tabular}%
}
\label{table:stats}
\end{table*}
\footnotetext{\scriptsize\url{https://en.wikipedia.org/wiki/List_of_Wikipedias}}  %

\begin{table}[t]
\caption{\textbf{Number of samples per language} in our multilingual data. ISO 639-1 language code is denoted as ``Code''.}
\centering
\begin{tabular}{crrr} \toprule
\textbf{Language (Code)} & \multicolumn{1}{c}{\textbf{\#Train}} & \multicolumn{1}{l}{\textbf{\#Val}} & \multicolumn{1}{l}{\textbf{\#Test}} \\ \toprule
English (EN) & 6,403,095 & 500 & 500 \\
Chinese (ZH) & 1,248,720 & 500 & 500 \\
Japanese (JA) & 1,293,628 & 500 & 500 \\
Spanish (ES) & 1,712,046 & 500 & 500 \\
French (FR) & 2,409,552 & 500 & 500 \\
Portuguese (PT) & 1,087,824 & 500 & 500 \\
Italian (IT) & 1,747,412 & 500 & 500 \\
German (DE) & 2,681,896 & 500 & 500 \\ \midrule
\textbf{Total} & 18,584,173 & 4,000 & 4,000 \\ \bottomrule
\end{tabular}
\label{table:NumSamplesPerLang}
\end{table}

\subsubsection{Pretraining}
In pretraining BROS$_{\text{BASE}}$, we used 6.4 million English samples generated by Webvicob and trained the model for 5 epochs. The training procedure involved randomly selecting 512 consecutive tokens from each document, similar to LayoutLMv2~\cite{xu2021layoutlmv2}. The remaining training hyperparameters were kept the same as BROS$_{\text{BASE}}$.

As can be seen in Table~\ref{table:stats} and Table~\ref{table:NumSamplesPerLang}, we pretrained LayoutXLM$_{\text{BASE}}$ using 18.6 million multilingual data generated by Webvicob, which included eight languages: English, Chinese, Japanese, Spanish, French, Portuguese, Italian, and German. The model was pretrained for 5 epochs with a batch size of 64, following the procedure described in LayoutXLM. In line with the multilingual sampling strategy~\cite{XLM,InfoXLM,layoutxlm}, each batch was sampled based on the language's frequency of occurrence in the data, with the probability $p_l \propto (\frac{N_l}{N})^{\alpha}$, where $N$ is the total number of data and $N_l$ is the number of data for language $l$. The parameter $\alpha$ was set to 0.7 to account for imbalanced data distribution.

The open-sourced Donut$_{\text{Proto}}$ was pretrained with 1.2 million samples generated by SynthDoG, including 400,000 samples each for Korean, Japanese, and English. The pretraining was performed for a duration of 5 days using 8 NVIDIA V100 GPUs, totaling 40 GPU days, with a batch size of 8. A similar approach was taken, where 400,000 samples each for Korean, Japanese, and English were generated by Webvicob and Donut$_{\text{Proto}}$ was trained for 3 days on 8 NVIDIA V100 GPUs (24 GPU days) with a batch size of 16. The images are cropped and then resized to a size of 2,048 in width and 1,536 in height for training purposes.

\subsubsection{Finetuning}
We finetune BROS with DocVQA datasets for 16K iterations. 64 batch size, Adam~\cite{Adamoptim} optimizer, 5e-5 learning rate, and cosine annealing learning rate scheduler~\cite{CosineLR} are adapted.

Since the exact finetuning schedule has not been disclosed, we set up a ``rough'' finetuning schedule and compared results. For XFUND Semantic Entity Recognition (SER), we finetune LayoutXLM with 10K iterations, 64 batch-size, AdamW~\cite{AdamWoptim} optimizer, 1e-4 learning rate, linear decay learning rate, and warmup~\cite{warmup} learning rate for 10\% of total iterations. We use unilm library\footnote{\scriptsize\url{https://github.com/microsoft/unilm}} for LayoutXLM finetuning experiments.

Donut$_{\text{Proto}}$ was finetuned for 300 epochs for the DocVQA tasks with an image size of 2,560 width and 2,048 height. For FUNSD and XFUND tasks, we finetuned the model for 2,000 epochs using images with a width of 2,048 and a height of 1,536. For all downstream tasks, 8 batch size, Adam~\cite{Adamoptim} optimizer, 1.5e-5 learning rate, and cosine annealing learning rate scheduler are adapted.

\subsection{Experimental Results}
\subsubsection{BROS}

The corpus of DocVQA Task 1 and IIT-CDIP are very similar, while Webvicob-generated data has more various corpus (will be shown in Section~\ref{sec:pca_analysis}).

The results also reflect this pattern. The results of Table~\ref{table:docvqa_results} are categorized based on the quantity of data used in pretraining. Because of the significant amount of data specific to a particular domain within IIT-CDIP, it is possible to build a domain-specialized model. On the other hand, the extensive diversity of domains represented in Webvicob-generated data resulted in the development of a model that excels in the DocVQA Task 3.

 The use of IIT-CDIP data for pretraining resulted in favorable performance on DocVQA Task 1. Conversely, when Webvicob data was used, DocVQA Task 3 performed much better. Surprisingly, the performance of Task 3 using Webvicob 1M has better performance than BROS using IIT-CDIP 11M. When a combination of both IIT-CDIP and Webvicob-generated data were used in equal portions for pretraining, there was a moderate enhancement in performance for both tasks.

\begin{table}[t]

\caption{\textbf{Quantitative comparison in DocVQA tasks} with BROS$_{\text{BASE}}$ according to various settings of pretraining corpus. For an apples-to-apples comparison, we control the amount of pretraining data in experiments. The score is measured with Average Normalized Levenshtein Similarity (ANLS).}
\begin{center}
\tabcolsep=0.1cm
\begin{tabular}{lcccc}
\toprule
\textbf{Model} & \textbf{IIT-CDIP} & \textbf{Webvicob} & \textbf{Task 1}  &  \textbf{Task 3}  \\
\toprule
BROS & - & - & 62.98 &  25.22 \\ \midrule
BROS & 500K & - & 65.01 & 25.12 \\
BROS & - & 500K & 64.62 & 26.20 \\
BROS & 250K & 250K & 65.56 & 26.56 \\ \midrule
BROS & 1M & - & 65.27 & 26.31 \\
BROS & - & 1M & 64.92 & 28.09 \\
BROS & 500K & 500K & 65.67 & 26.51 \\ \midrule
BROS & 11M & - & 68.07 & 24.76 \\
BROS & - & 6.4M & 65.63 & 27.85 \\
\bottomrule
\end{tabular}%
\label{table:docvqa_results}
\end{center}
\end{table}

\subsubsection{LayoutXLM}\label{sec:multi}
We report the f1 scores of Semantic Entity Recognition (SER) for FUNSD and XFUND datasets. Table~\ref{table:XfundMultitaskFinetune} shows f1 scores of multitask finetuning with eight languages. Under the same finetuning setup, Webvicob data shows comparable performance. Please note that the total number of pretraining iterations of LayoutXLM (Webvicob) is $\sim$1.45M, much smaller than the iteration of LayoutXLM (PDF 22M + IIT 8M), which is $\sim$2.34M. 

\begin{table*}[t]
\caption{\textbf{Multitask finetuning Accuracy (F1) for Semantic Entity Recognition (SER) with FUNSD and XFUND datasets.} We perform multitask finetuning LayoutXLM$_{\text{BASE}}$ under same setting with eight languages. Despite using only approximately 60\% of the training data and iterations, the ultimate mean performance being competitive.}
\resizebox{1.0\textwidth}{!}{%
\begin{tabular}{@{}lcccccccccc@{}}

\toprule
\textbf{Model} & \multicolumn{1}{c}{\textbf{FUNSD}} & \multicolumn{1}{c}{\textbf{ZH}} & \multicolumn{1}{c}{\textbf{JA}} & \multicolumn{1}{c}{\textbf{ES}} & \multicolumn{1}{c}{\textbf{FR}} & \multicolumn{1}{c}{\textbf{IT}} & \multicolumn{1}{c}{\textbf{DE}} & \multicolumn{1}{c}{\textbf{PT}} & \multicolumn{1}{c}{\textbf{Avg.}} \\
\toprule
LayoutXLM (PDF 22M + IIT 8M) & 0.785 & 0.897 & 0.787 & 0.744 & 0.791 & 0.811 & 0.832 & 0.824 & 0.785 \\
LayoutXLM (Webvicob 18.6M) & 0.727 & 0.893 & 0.794 & 0.686 & 0.749 & 0.765 & 0.760 & 0.764 & 0.767 \\ \bottomrule

\end{tabular}%
}

\label{table:XfundMultitaskFinetune}
\end{table*}

\subsubsection{Donut}
We report finetuning results of Donut$_{\text{Proto}}$ on Table~\ref{table:donut_results}. Donut$_{\text{Proto}}$ trained using Webvicob data has achieved superior performance on all four tasks, despite 60\% GPU training days compared to the Donut$_{\text{Proto}}$ trained with SynthDoG-generated data. While SynthDoG also made use of the Wikipedia corpus, the outstanding performance of Webvicob-generated data is noteworthy. We speculate that there are two reasons for this. First, the presence of text and images that are contextually relevant in Webvicob helps to accurately reflect the relationship between images and text. Second, the use of various types of real documents, such as tables, in the learning process can implicitly improve the learning of semantic information, such as key-value pairing.

\begin{table*}[t]
\caption{\textbf{Finetuning results of Donut$_{\text{Proto}}$.} The XFUND (JA) dataset is denoted as ``JA'', while DocVQA Task 1 and DocVQA Task 3 are denoted as ``Task 1'' and ``Task 3'' respectively. The evaluation of the Entity Recognition task in FUNSD and XFUND was conducted using TED-based accuracy. ANLS scores are used for the DocVQA tasks. Despite utilizing only 60\% of GPU days, the Donut$_{\text{Proto}}$ model trained on Webvicob-generated data exhibited better performance than the Donut$_{\text{Proto}}$ model trained on SynthDoG-generated data.}
\begin{center}
\tabcolsep=0.1cm

\begin{tabular}{lcccc}
\toprule
\multicolumn{1}{l}{\textbf{Model}} & \multicolumn{1}{c}{\textbf{FUNSD}} & \multicolumn{1}{c}{\textbf{JA}} & \multicolumn{1}{c}{\textbf{Task 1}} & \multicolumn{1}{c}{\textbf{Task 3}} \\
\toprule
Donut (Not Pretrained) & 0.1390 & 0.1942 & 0.1391 & 0.0877 \\
Donut (SynthDoG 1.2M) & 0.6059 & 0.7817 & 0.4477 & 0.1018 \\
Donut (Webvicob 1.2M) & 0.7397 & 0.8455 & 0.5607 & 0.1842 \\
Donut (SynthDoG 0.6M + Webvicob 0.6M) & 0.7576 & 0.8488 & 0.5622 & 0.2039 \\
Donut (SynthDoG 1.2M + Webvicob 1.2M) & 0.7738 & 0.8611 & 0.5636 & 0.2165 \\ \bottomrule
\end{tabular}%
\end{center}

\label{table:donut_results}

\end{table*}

\section{Discussion}
\subsection{Scale-up using CommonCrawl}
Our initial goal was to handle tons of HTML data, such as CommonCrawl dataset\footnote{\scriptsize\url{https://commoncrawl.github.io/cc-crawl-statistics/plots/crawlsize}} which consists of petabytes of data.

However, the massive variety in the HTML format made it difficult to design an integrated solution.
Currently, Webvicob is only specialized in a Wikipedia format.
Expanding the scope could be future work.  

\subsection{Webvicob with Augraphy}
\begin{figure}[htp]
    \captionsetup[subfigure]{oneside,margin={0.7cm,0cm}}
    
    \begin{subfigure}{0.235\textwidth}
        \includegraphics[width=\textwidth]{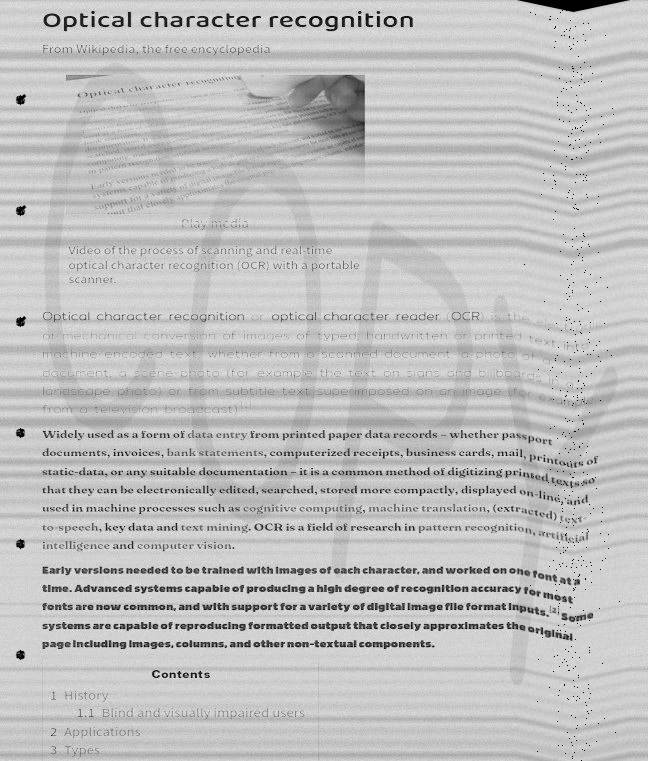}
    \end{subfigure}
    \begin{subfigure}{0.235\textwidth}
        \includegraphics[width=\textwidth]{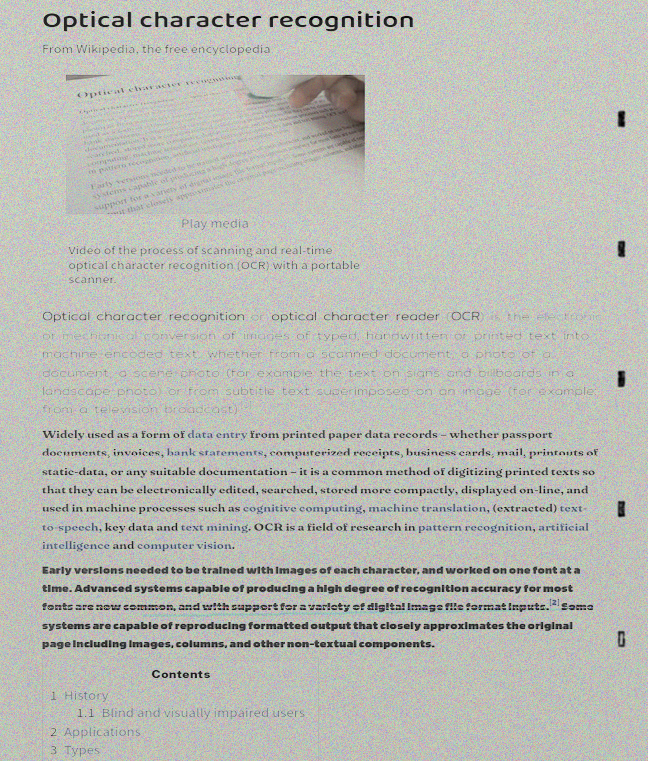}
    \end{subfigure}
    \begin{subfigure}{0.235\textwidth}
        \includegraphics[width=\textwidth]{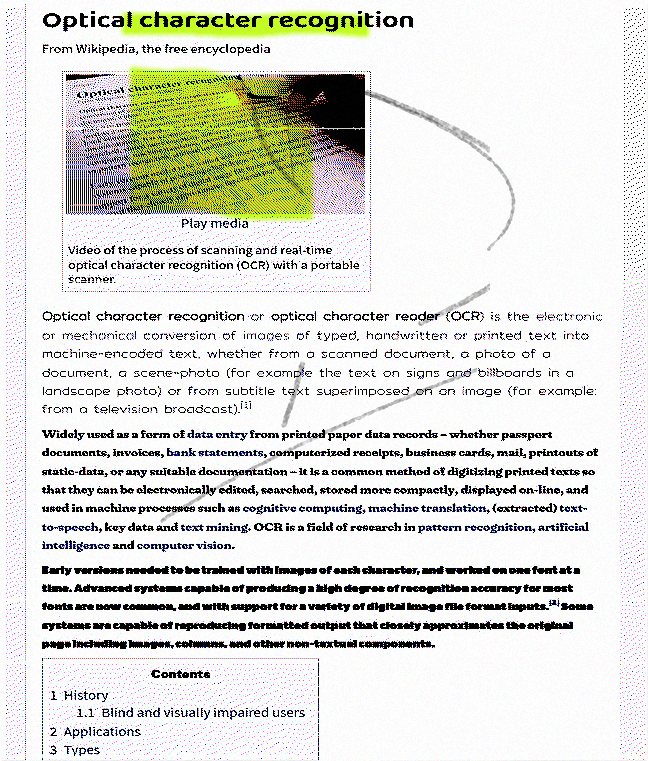}
    \end{subfigure}
    \begin{subfigure}{0.235\textwidth}
        \includegraphics[width=\textwidth]{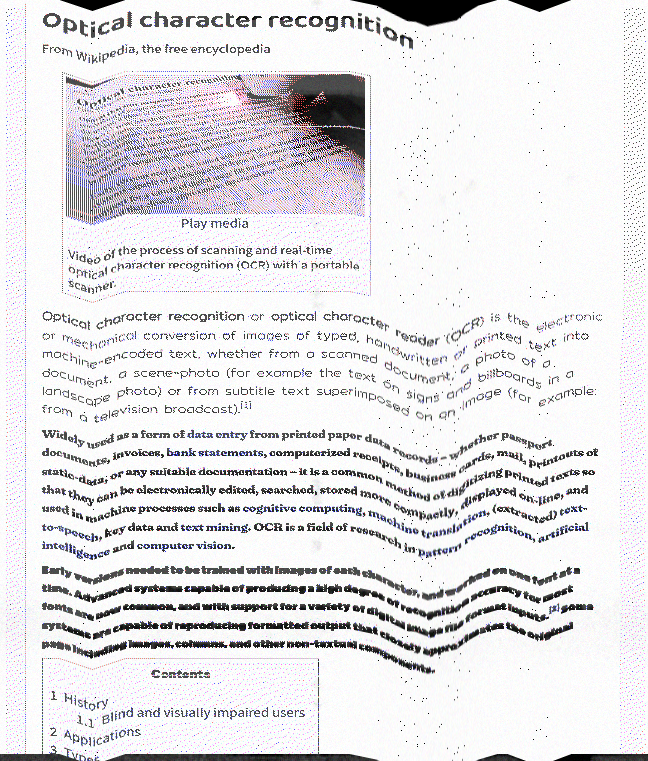}
    \end{subfigure}
    \hfill
    \caption{\textbf{Webvicob with Augraphy.} Augraphy has the capability to produce a pencil sketch or folding appearance.}
    \label{fig:augraphy2}
\end{figure}

We expect Webvicob to easily be integrated with any augmentation project, such as Augraphy~\cite{augraphy}, to boost the performance further. As can be seen in Figure~\ref{fig:augraphy2}, combining Webvicob and Augraphy can generate document images with rich visual effects.

\subsection{PCA analysis}
\label{sec:pca_analysis}
\begin{figure}[htp]
    \centering
    \captionsetup[subfigure]{oneside,margin={0.4cm,0cm},font=tiny}
    
    \begin{subfigure}{0.201\textwidth}
        \includegraphics[width=\textwidth]{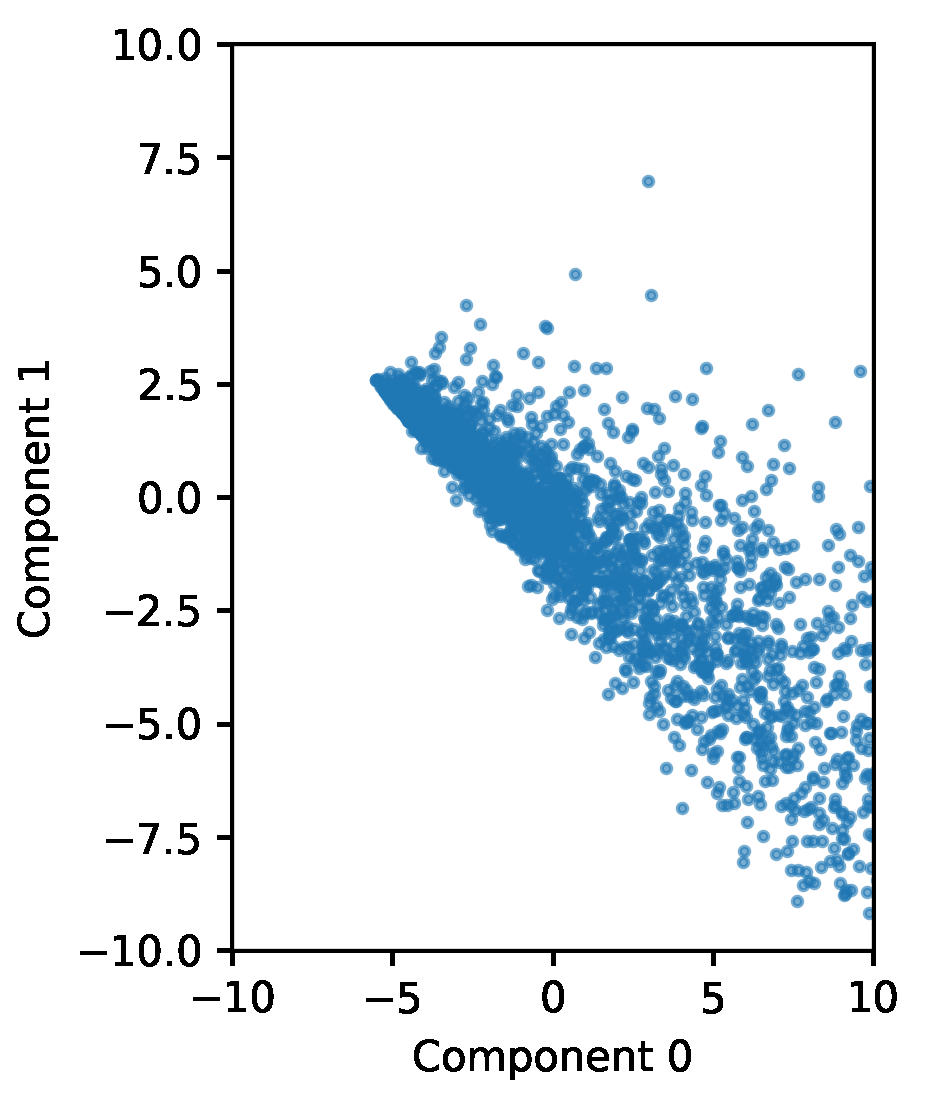}
    \end{subfigure}
    \begin{subfigure}{0.201\textwidth}
        \includegraphics[width=\textwidth]{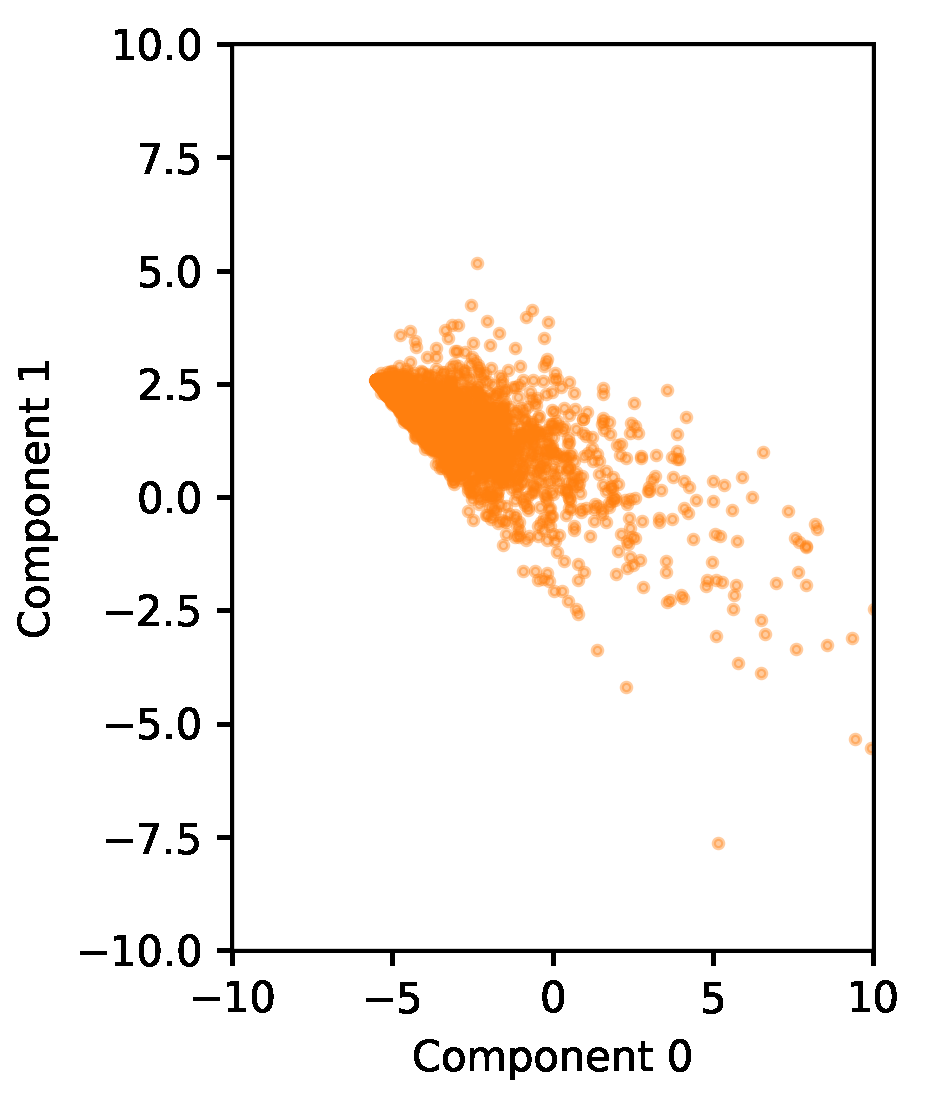}
    \end{subfigure}
    \begin{subfigure}{0.201\textwidth}
        \includegraphics[width=\textwidth]{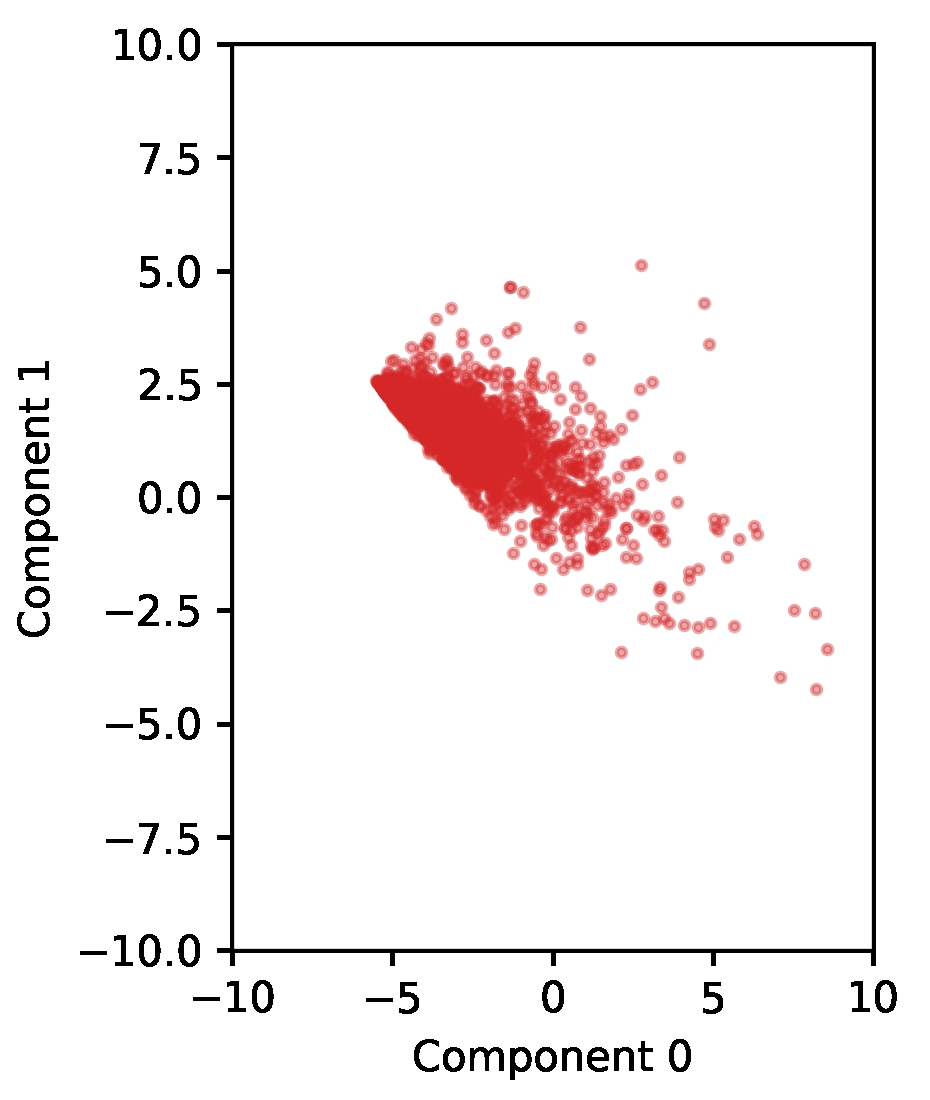}
    \end{subfigure}
    \begin{subfigure}{0.201\textwidth}
        \includegraphics[width=\textwidth]{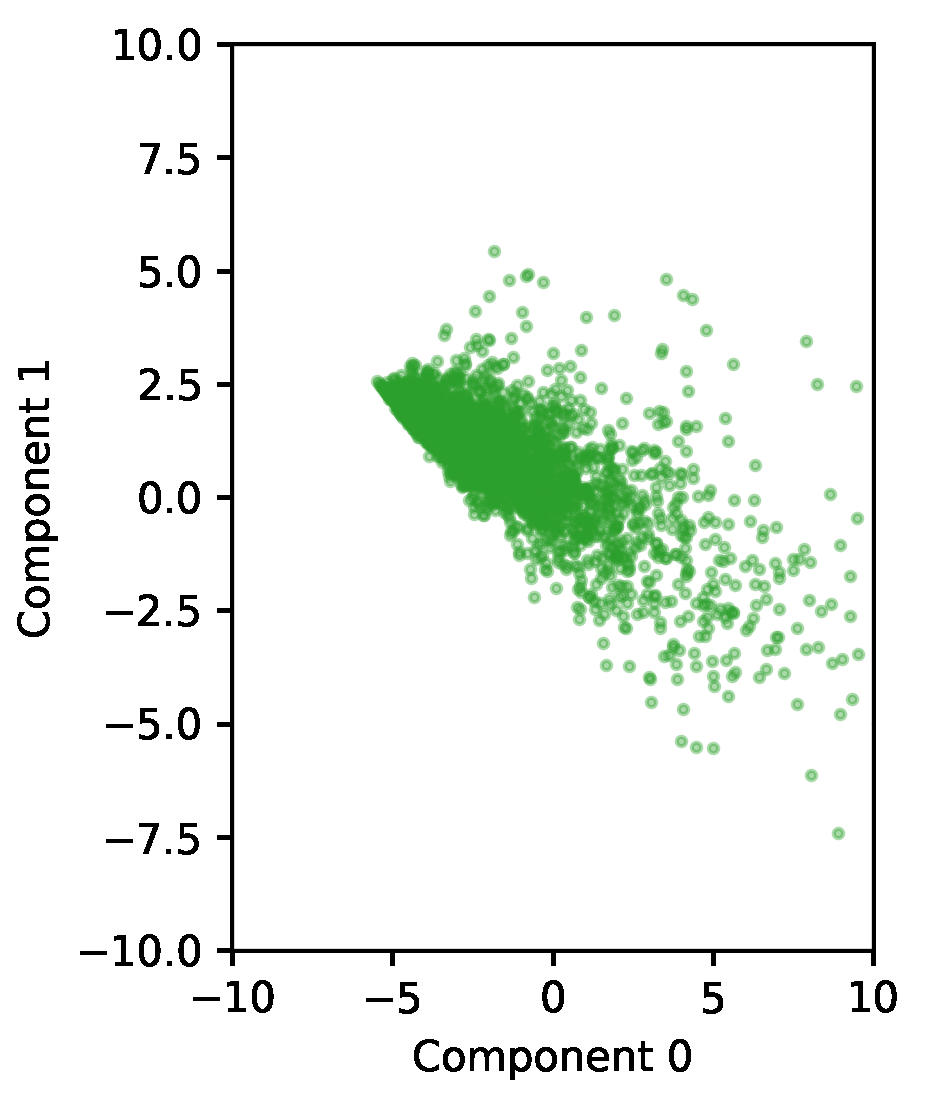}
    \end{subfigure}
    \hfill
    
    \begin{subfigure}{0.201\textwidth}
        \includegraphics[width=\textwidth]{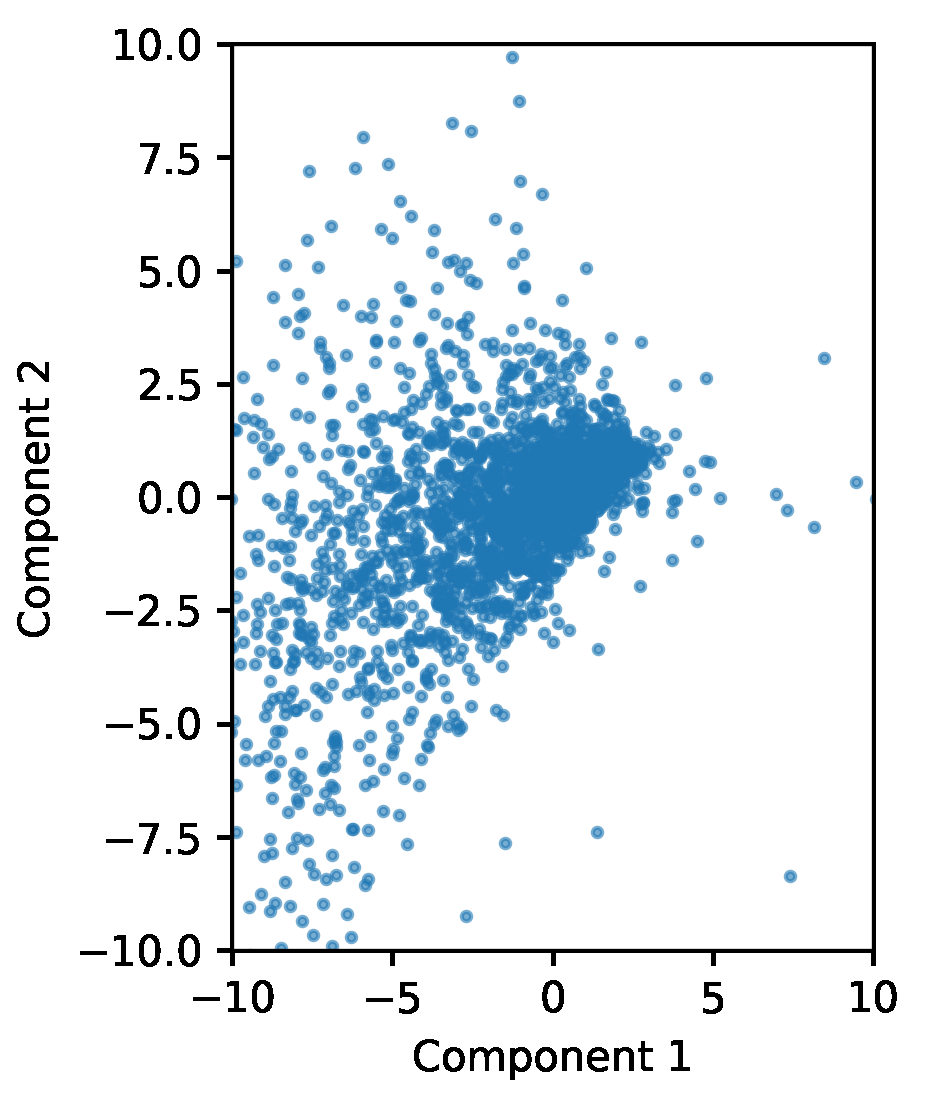}
    \end{subfigure}
    \begin{subfigure}{0.201\textwidth}
        \includegraphics[width=\textwidth]{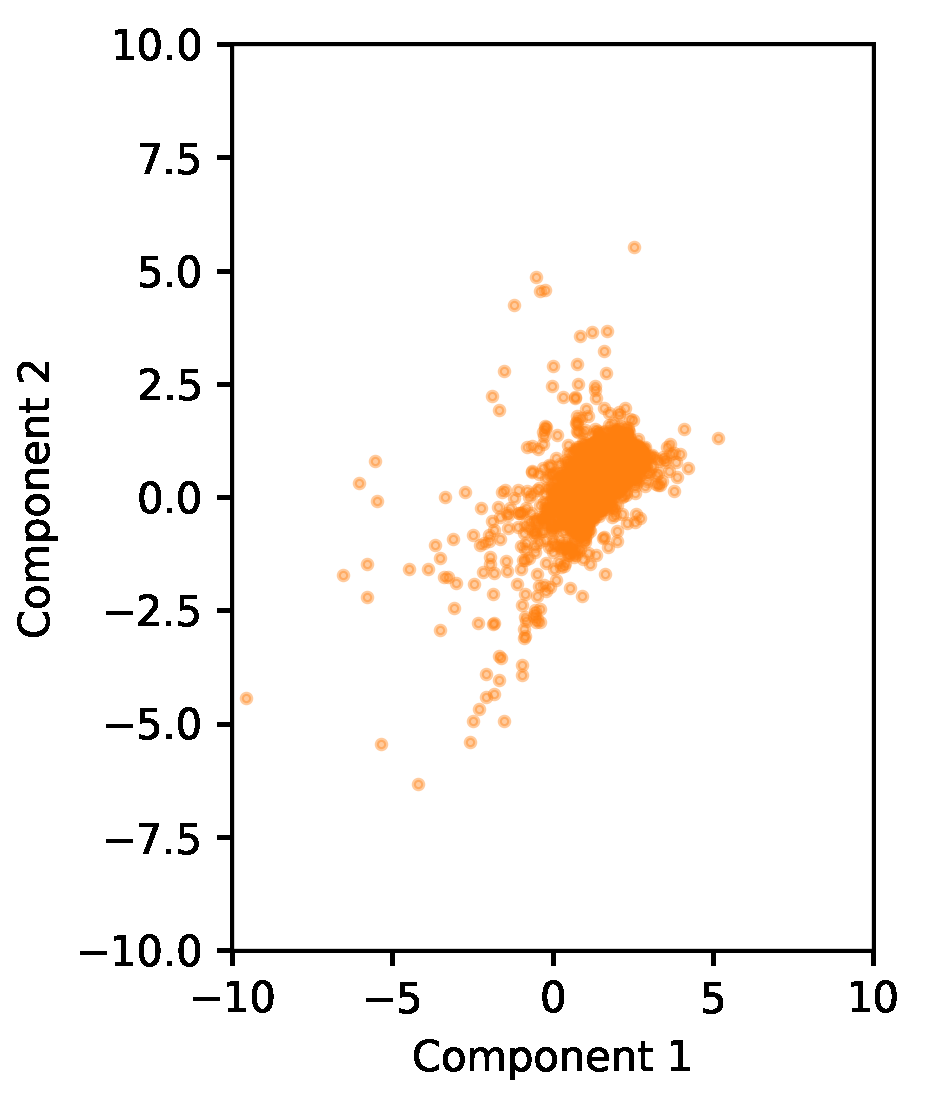}
    \end{subfigure}
    \begin{subfigure}{0.201\textwidth}
        \includegraphics[width=\textwidth]{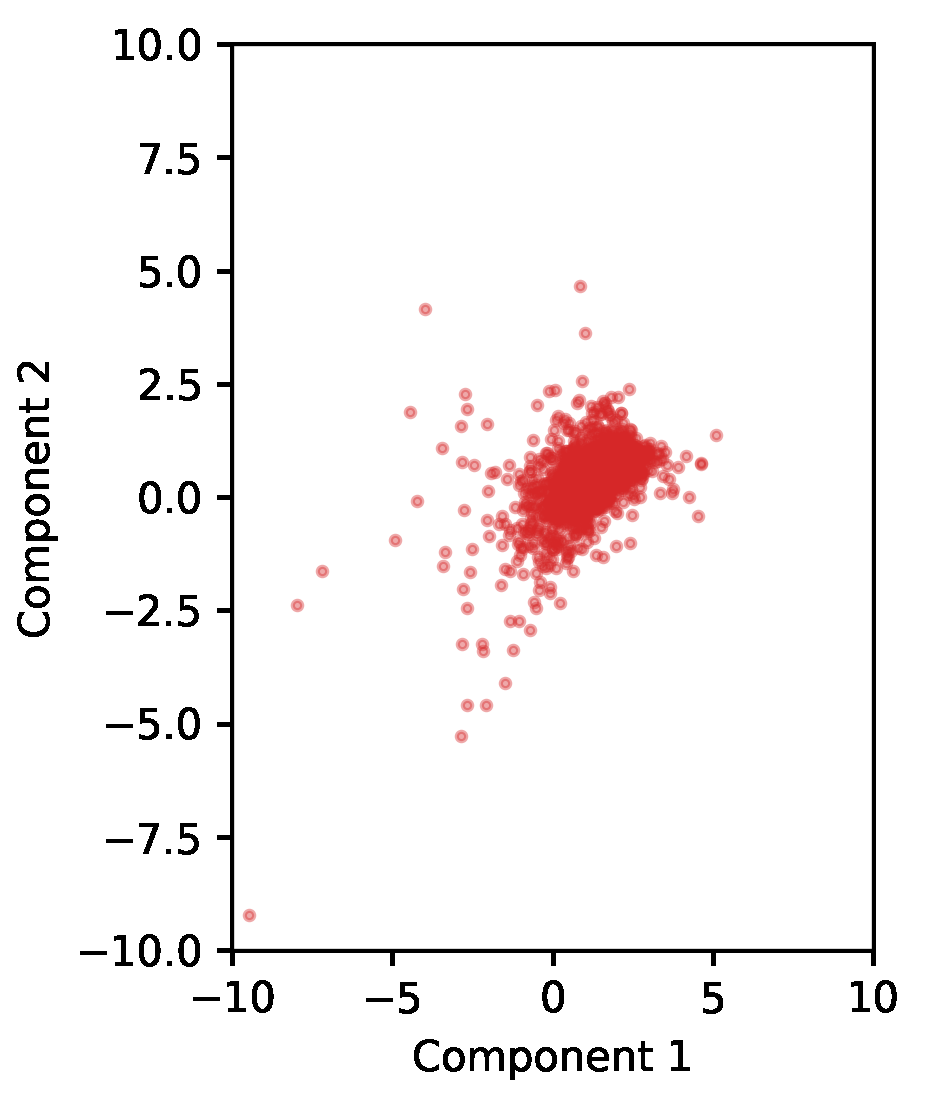}
    \end{subfigure}
    \begin{subfigure}{0.201\textwidth}
        \includegraphics[width=\textwidth]{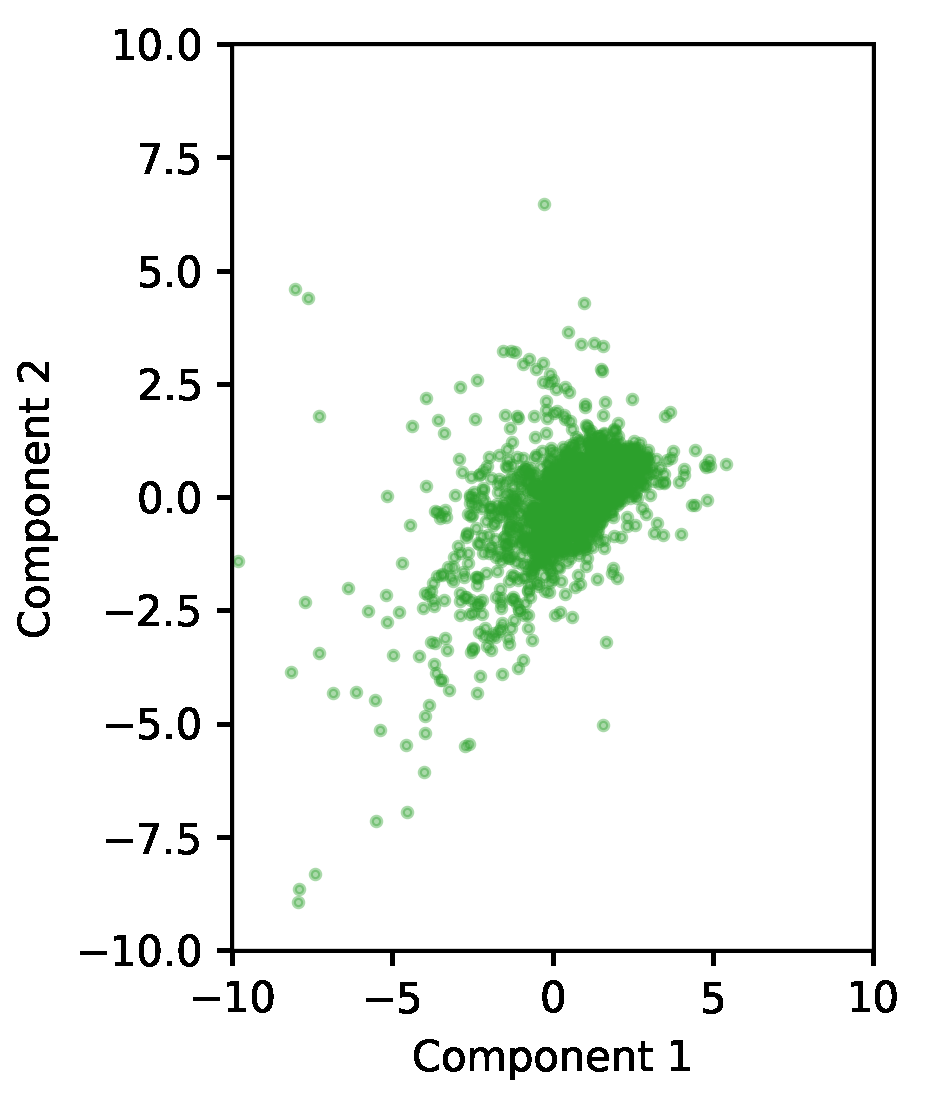}
    \end{subfigure}
    \hfill
    
    \begin{subfigure}{0.201\textwidth}
        \includegraphics[width=\textwidth]{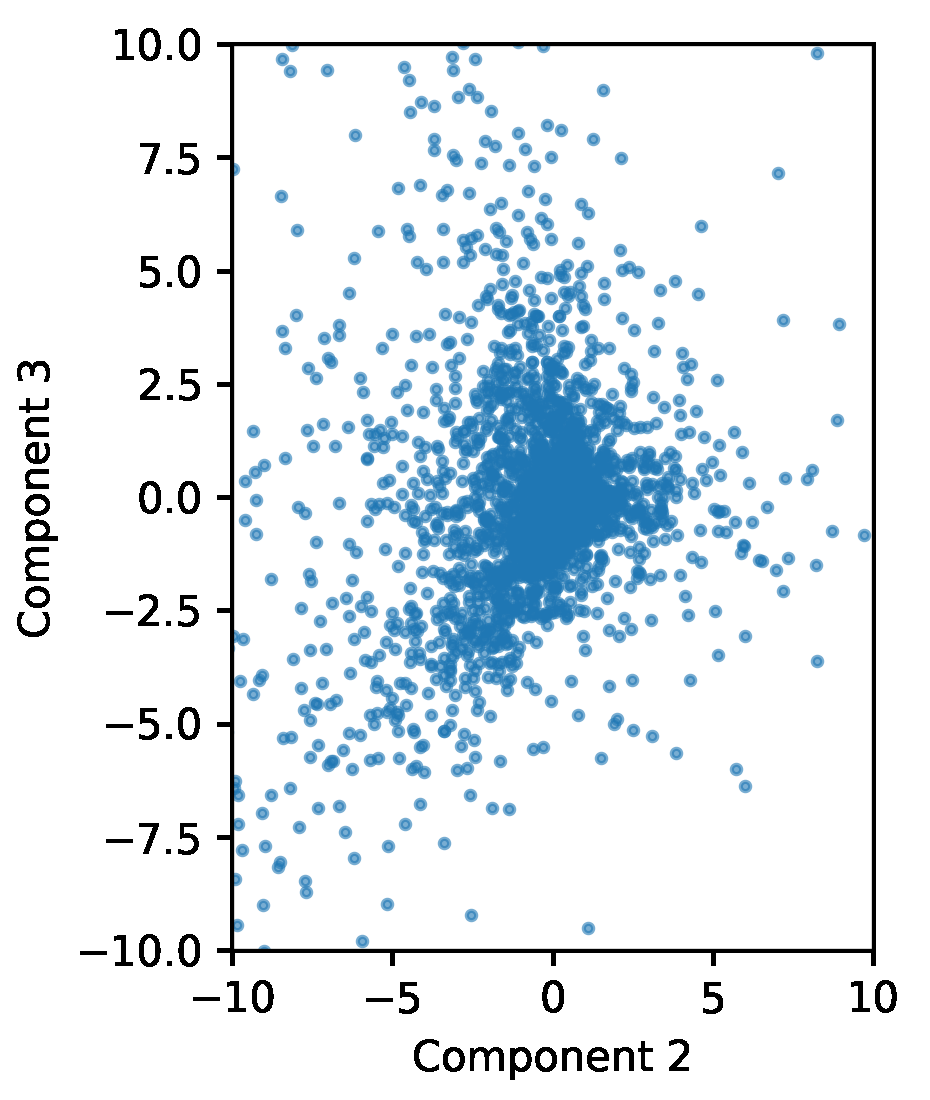}
    \end{subfigure}
    \begin{subfigure}{0.201\textwidth}
        \includegraphics[width=\textwidth]{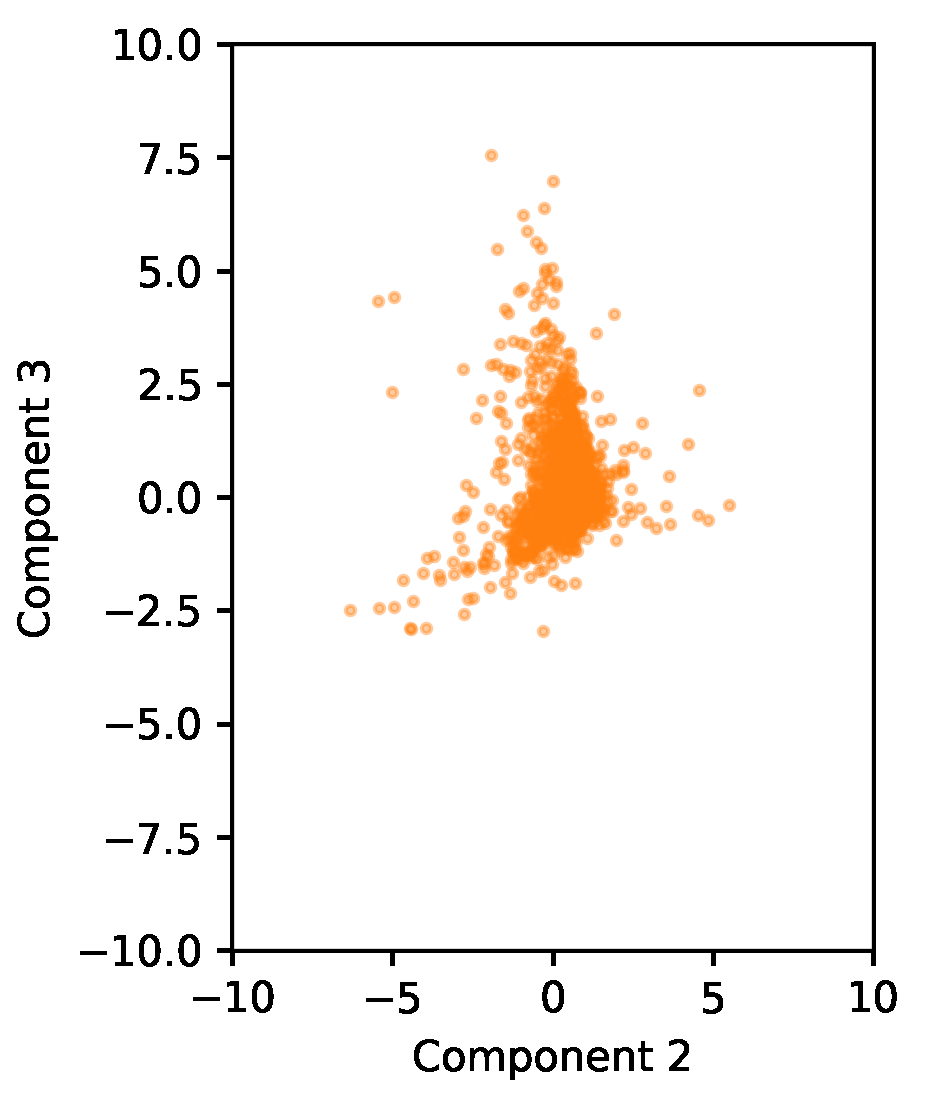}
    \end{subfigure}
    \begin{subfigure}{0.201\textwidth}
        \includegraphics[width=\textwidth]{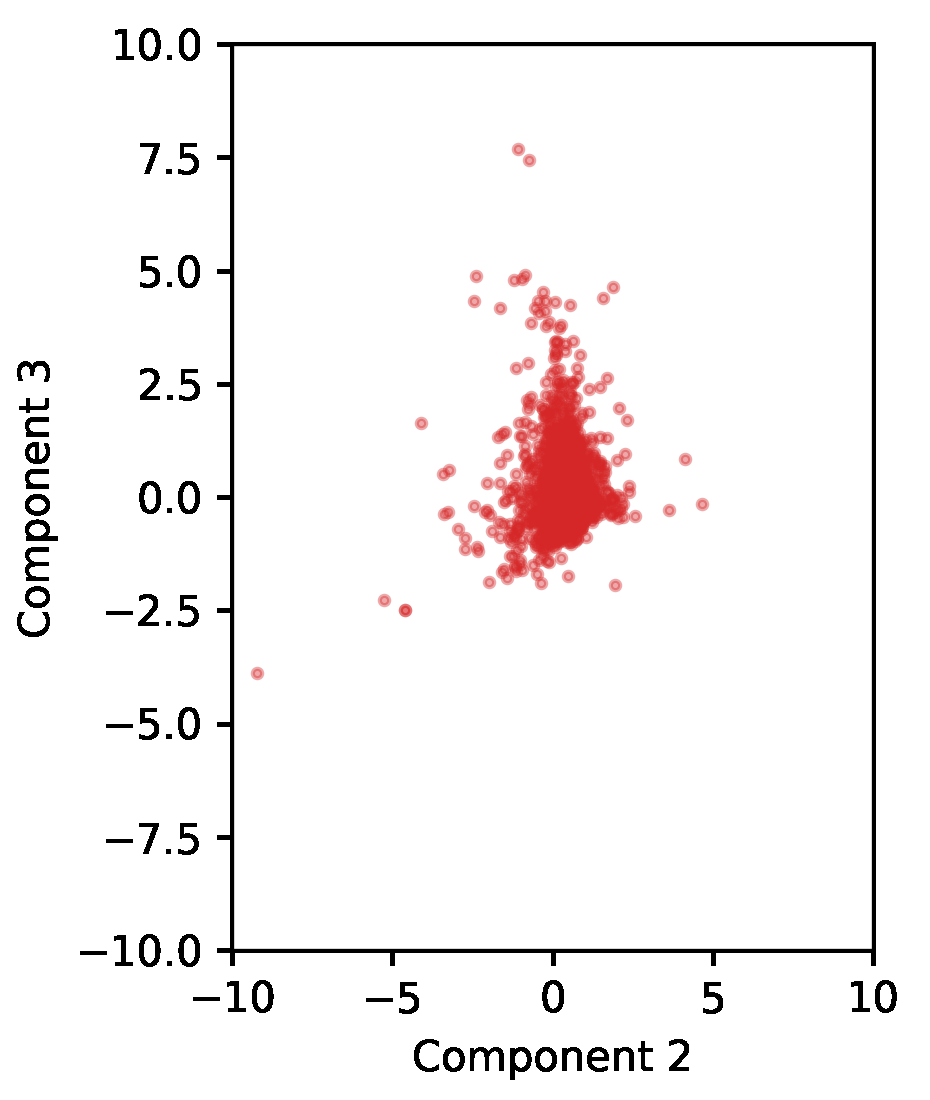}
    \end{subfigure}
    \begin{subfigure}{0.201\textwidth}
        \includegraphics[width=\textwidth]{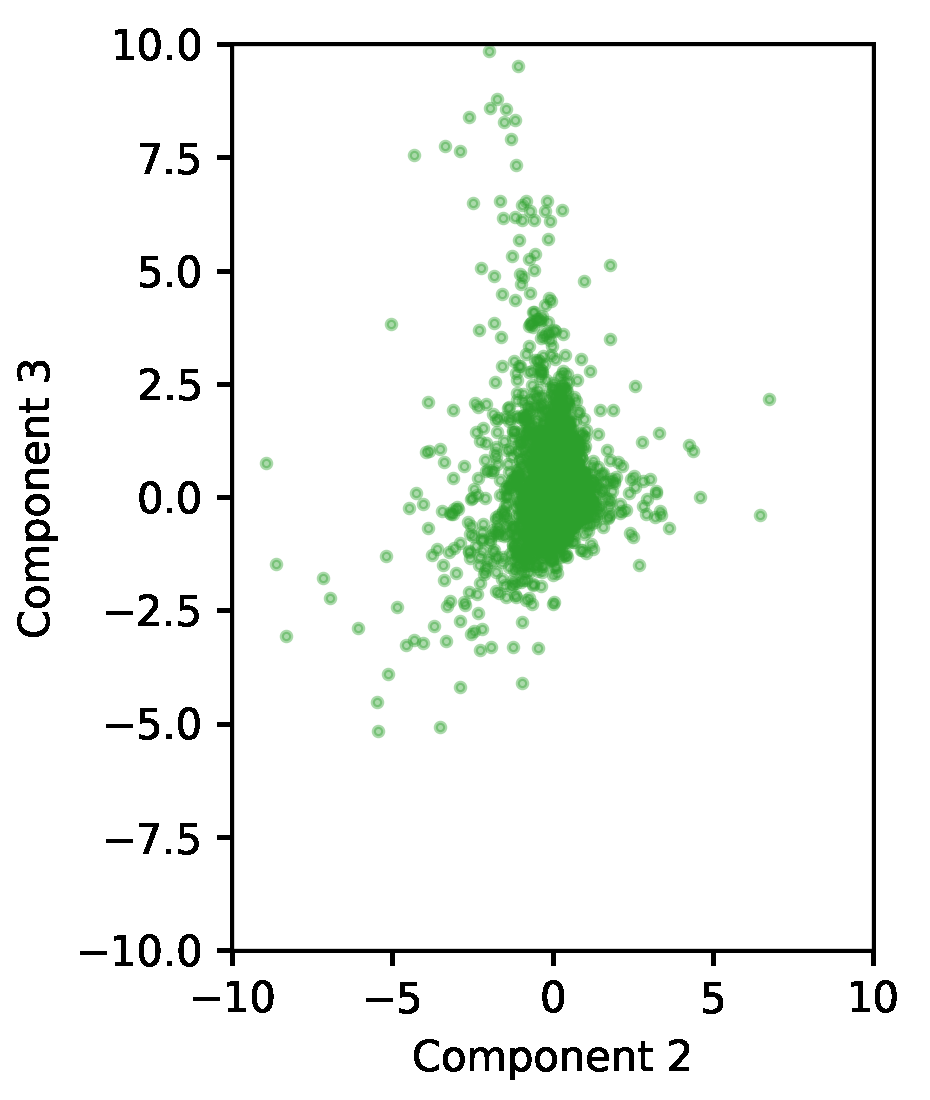}
    \end{subfigure}
    \hfill
    
    \begin{subfigure}{0.201\textwidth}
        \includegraphics[width=\textwidth]{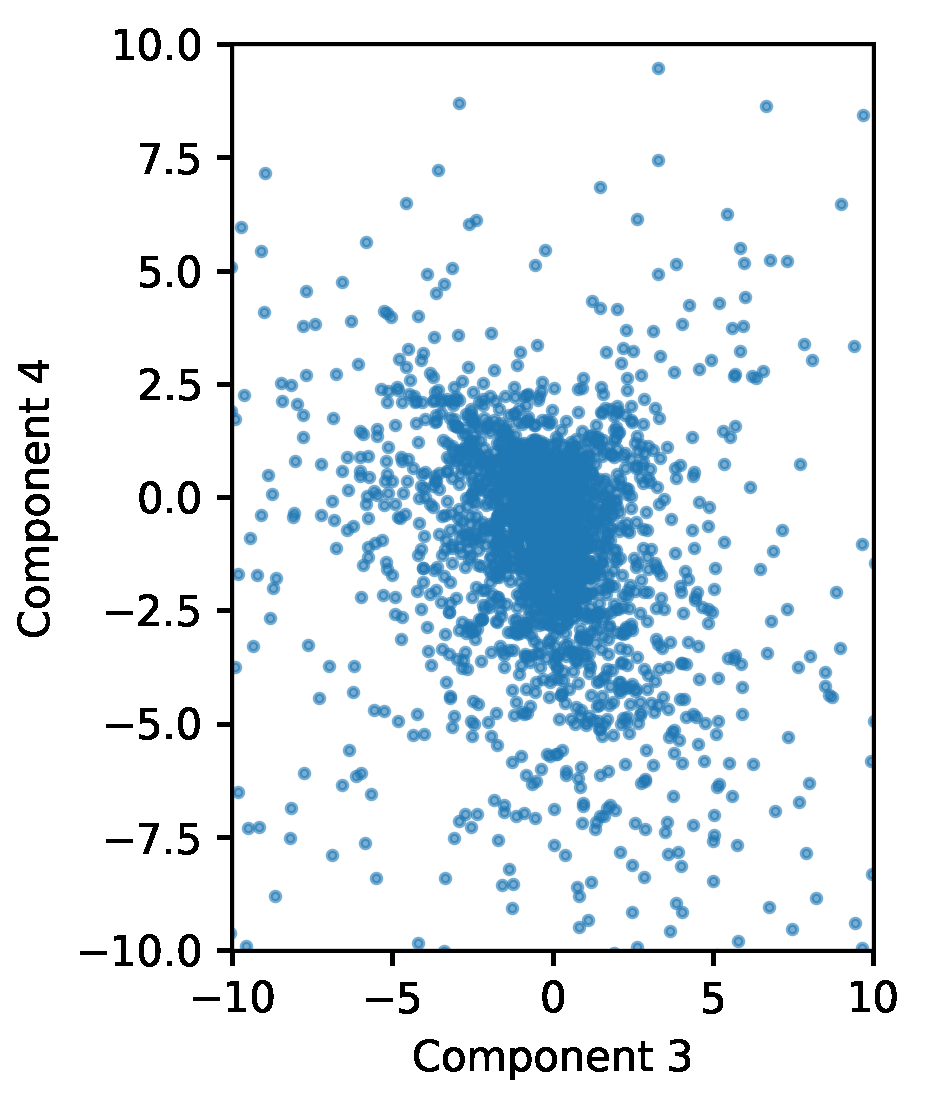}
        \caption{Webvicob}
    \end{subfigure}
    \begin{subfigure}{0.201\textwidth}
        \includegraphics[width=\textwidth]{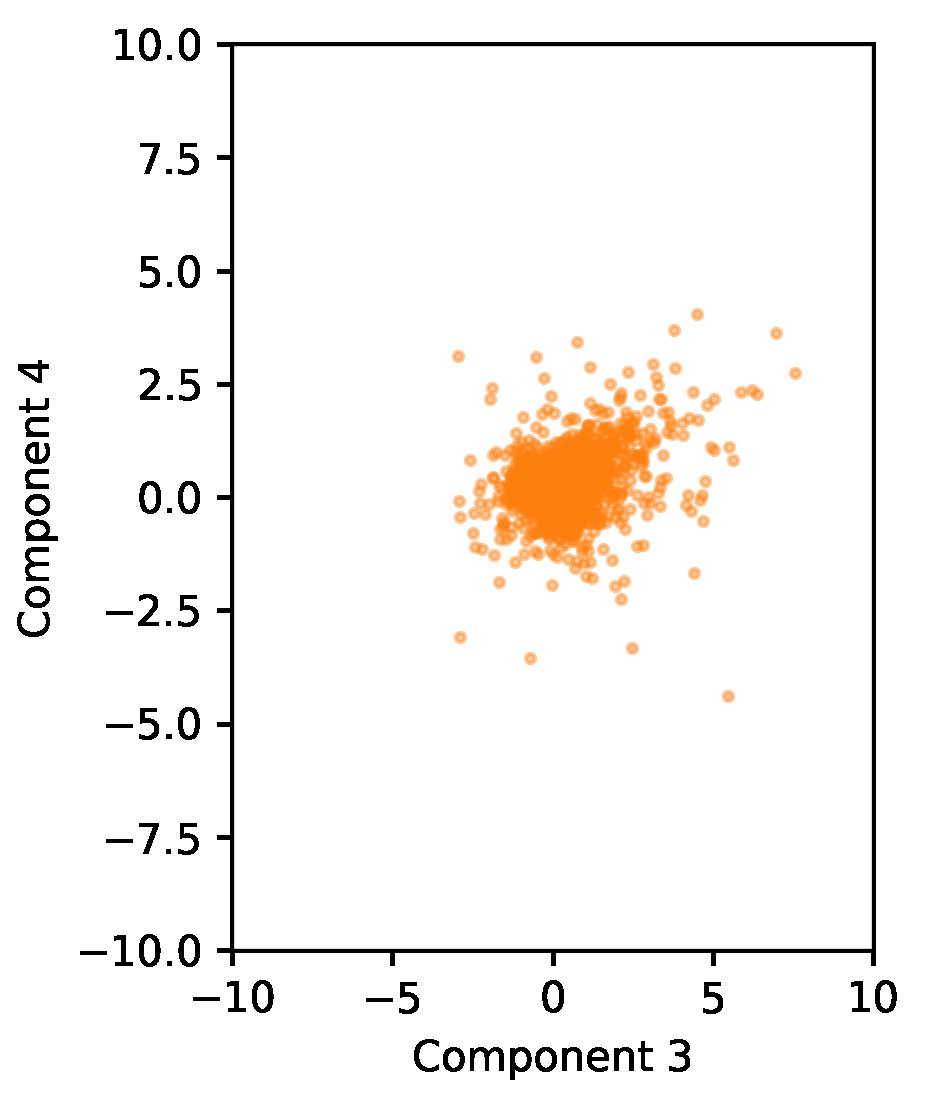}
        \caption{IIT-CDIP}
    \end{subfigure}
    \begin{subfigure}{0.201\textwidth}
        \includegraphics[width=\textwidth]{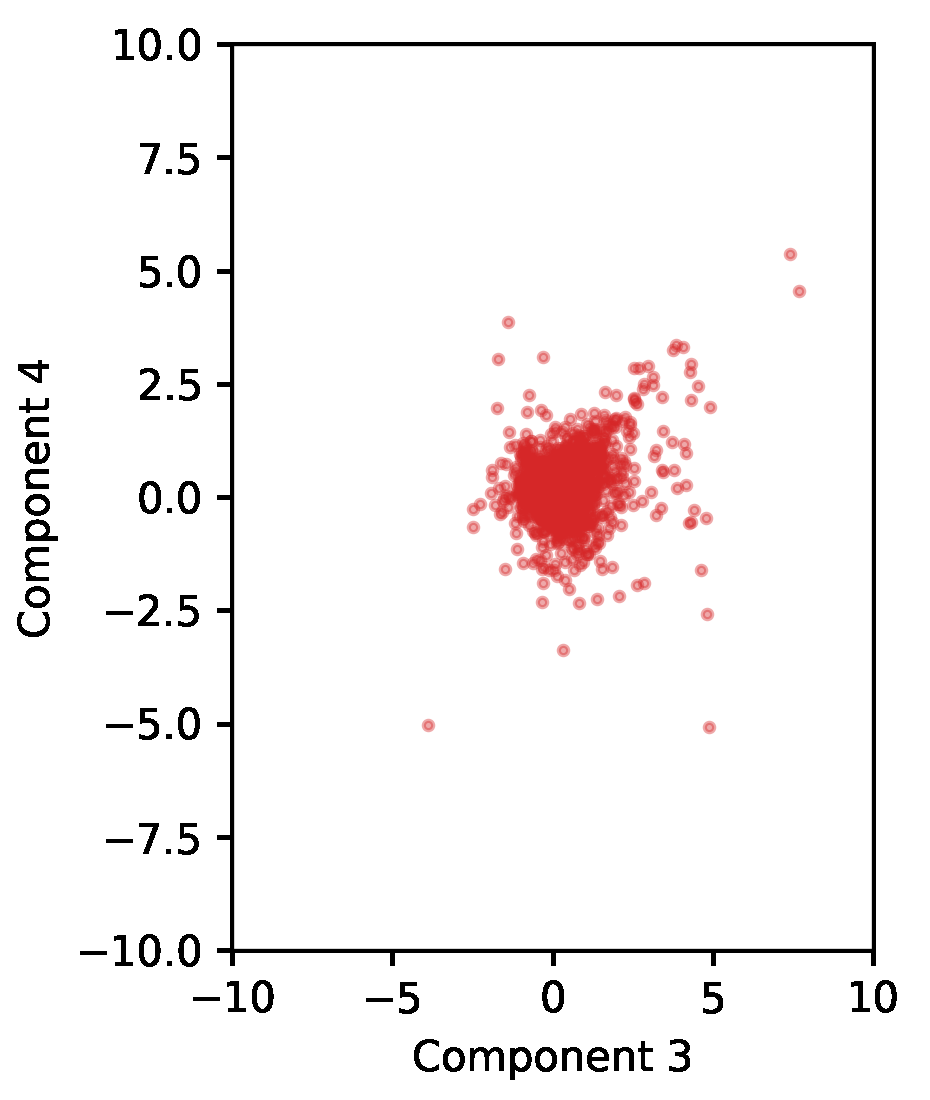}
        \caption{DocVQA Task 1}
    \end{subfigure}
    \begin{subfigure}{0.201\textwidth}
        \includegraphics[width=\textwidth]{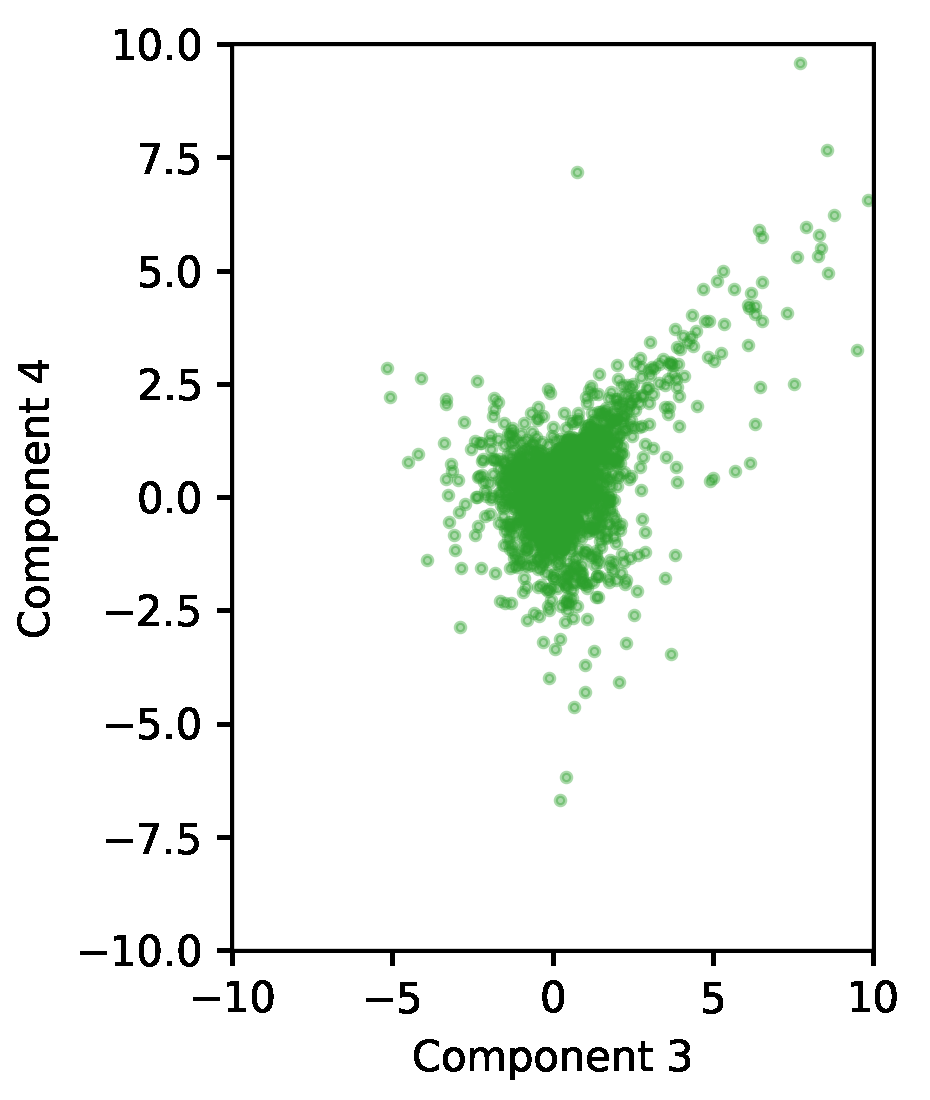}
        \caption{DocVQA Task 3}
    \end{subfigure}
    \hfill
    
    \caption{\textbf{Visualization of PCA Results.} Components 0, 1, 2, 3, 4}
    \label{fig:pca1}
\end{figure}

\begin{figure}[htp]
    \centering
    \captionsetup[subfigure]{oneside,margin={0.2cm,0cm},font=tiny}
    
    \begin{subfigure}{0.201\textwidth}
        \includegraphics[width=\textwidth]{figure/pca/4_5_Webvicob.png}
    \end{subfigure}
    \begin{subfigure}{0.201\textwidth}
        \includegraphics[width=\textwidth]{figure/pca/4_5_iit.png}
    \end{subfigure}
    \begin{subfigure}{0.201\textwidth}
        \includegraphics[width=\textwidth]{figure/pca/4_5_vqa_t1.png}
    \end{subfigure}
    \begin{subfigure}{0.201\textwidth}
        \includegraphics[width=\textwidth]{figure/pca/4_5_vqa_t3.png}
    \end{subfigure}
    \hfill
    
    \begin{subfigure}{0.201\textwidth}
        \includegraphics[width=\textwidth]{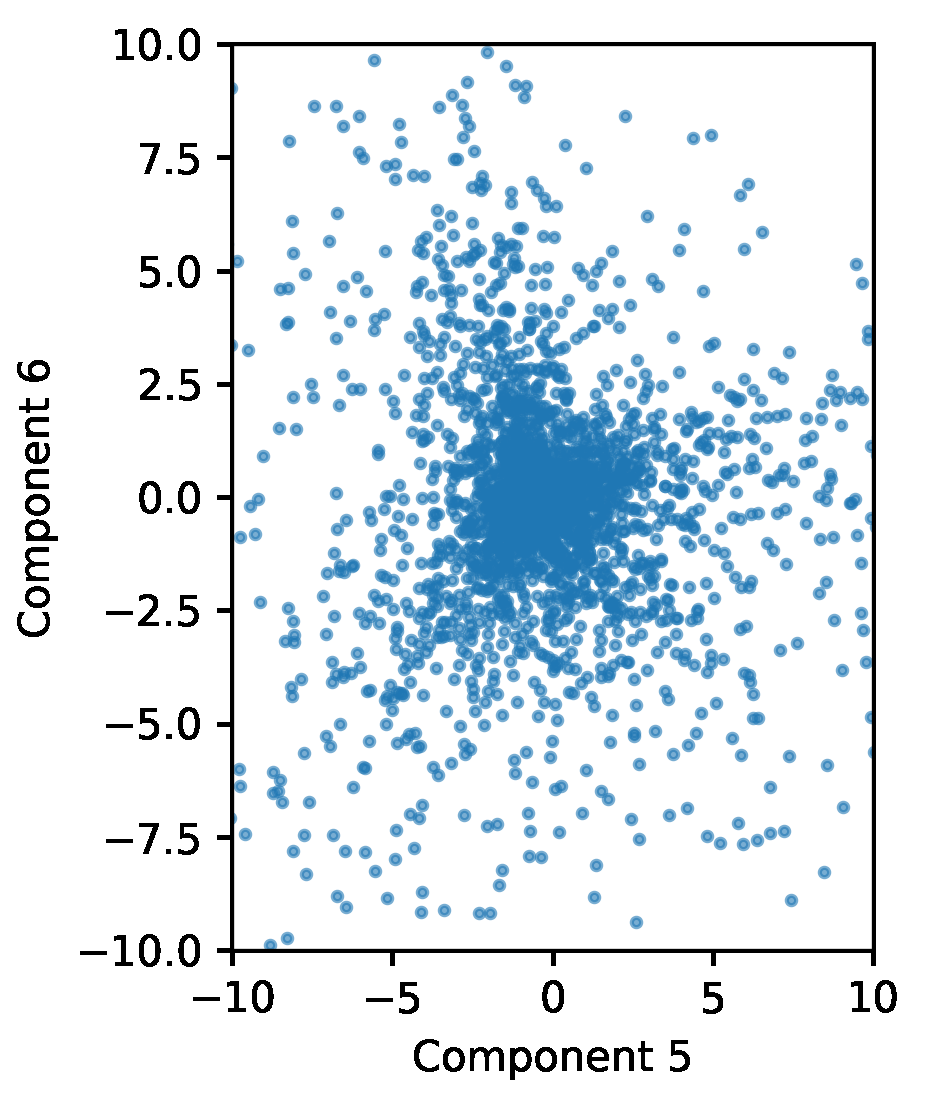}
    \end{subfigure}
    \begin{subfigure}{0.201\textwidth}
        \includegraphics[width=\textwidth]{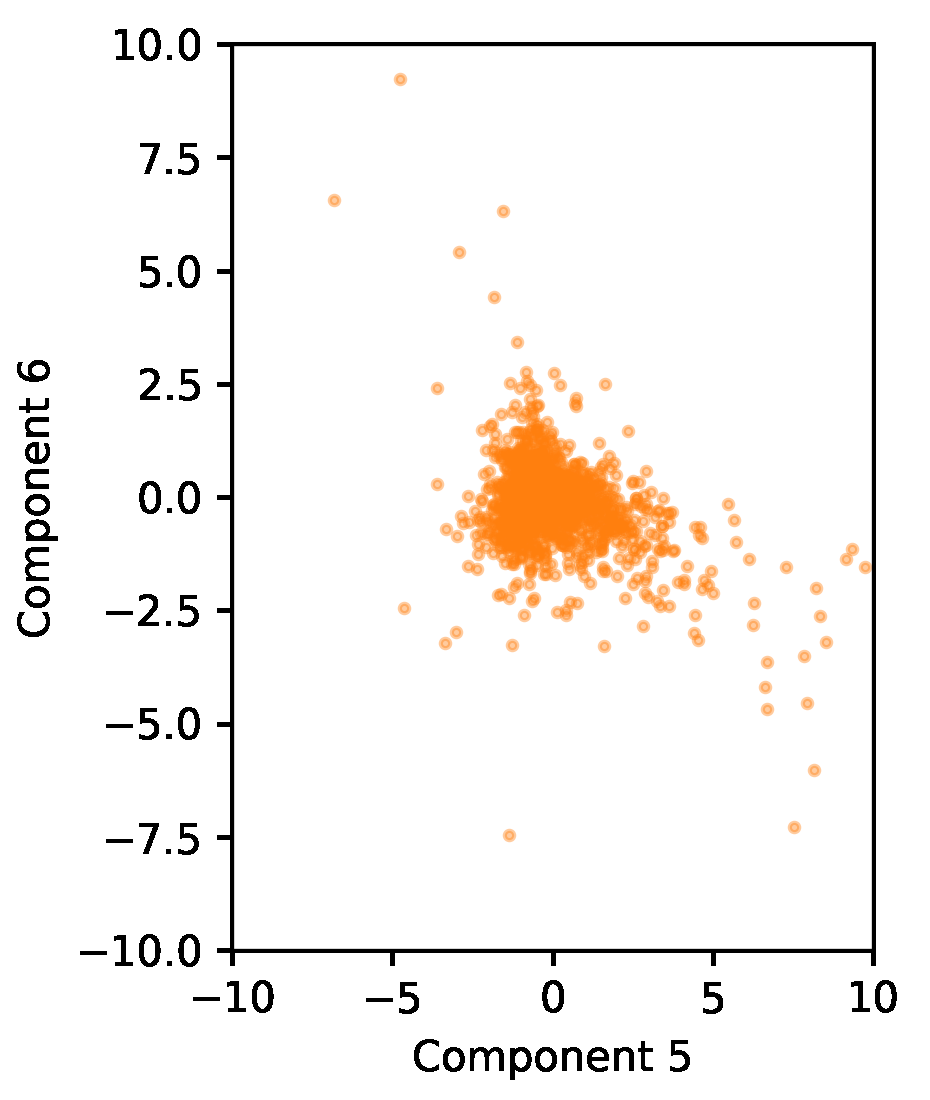}
    \end{subfigure}
    \begin{subfigure}{0.201\textwidth}
        \includegraphics[width=\textwidth]{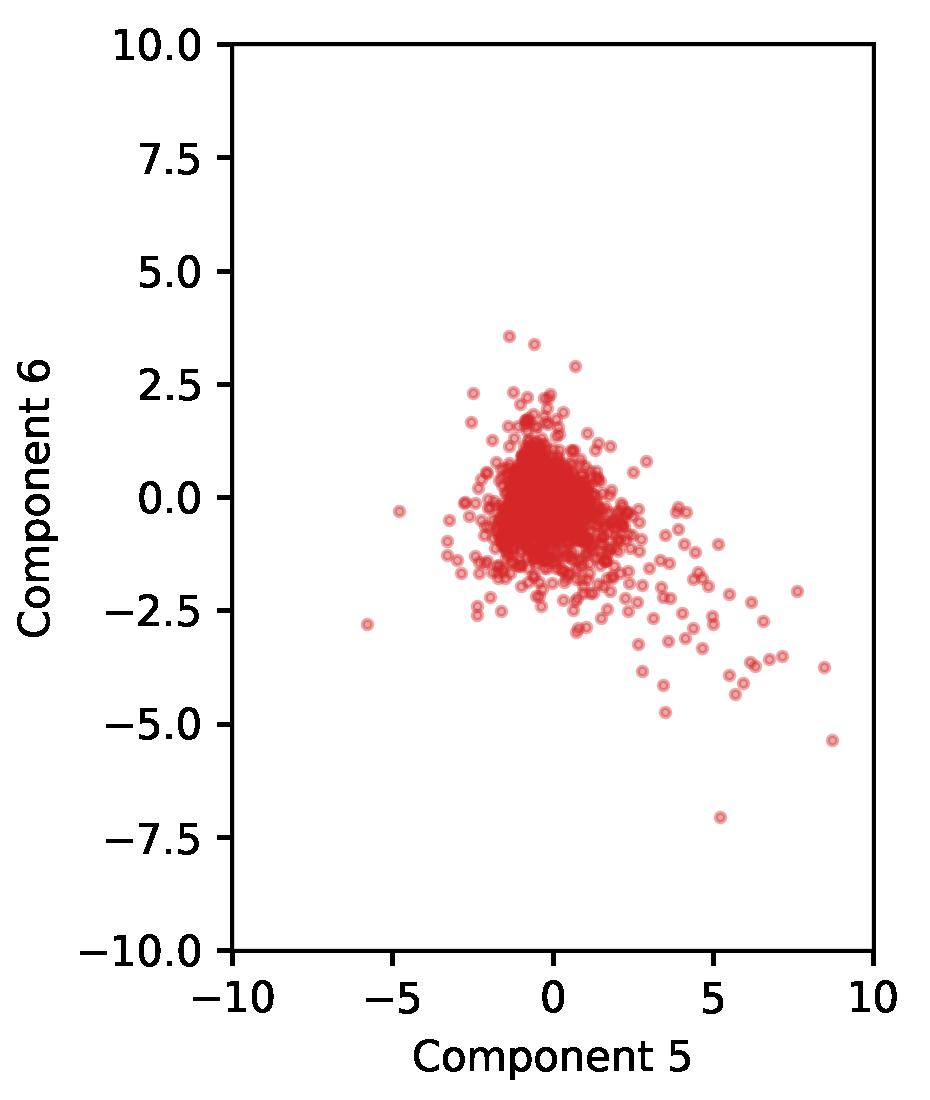}
    \end{subfigure}
    \begin{subfigure}{0.201\textwidth}
        \includegraphics[width=\textwidth]{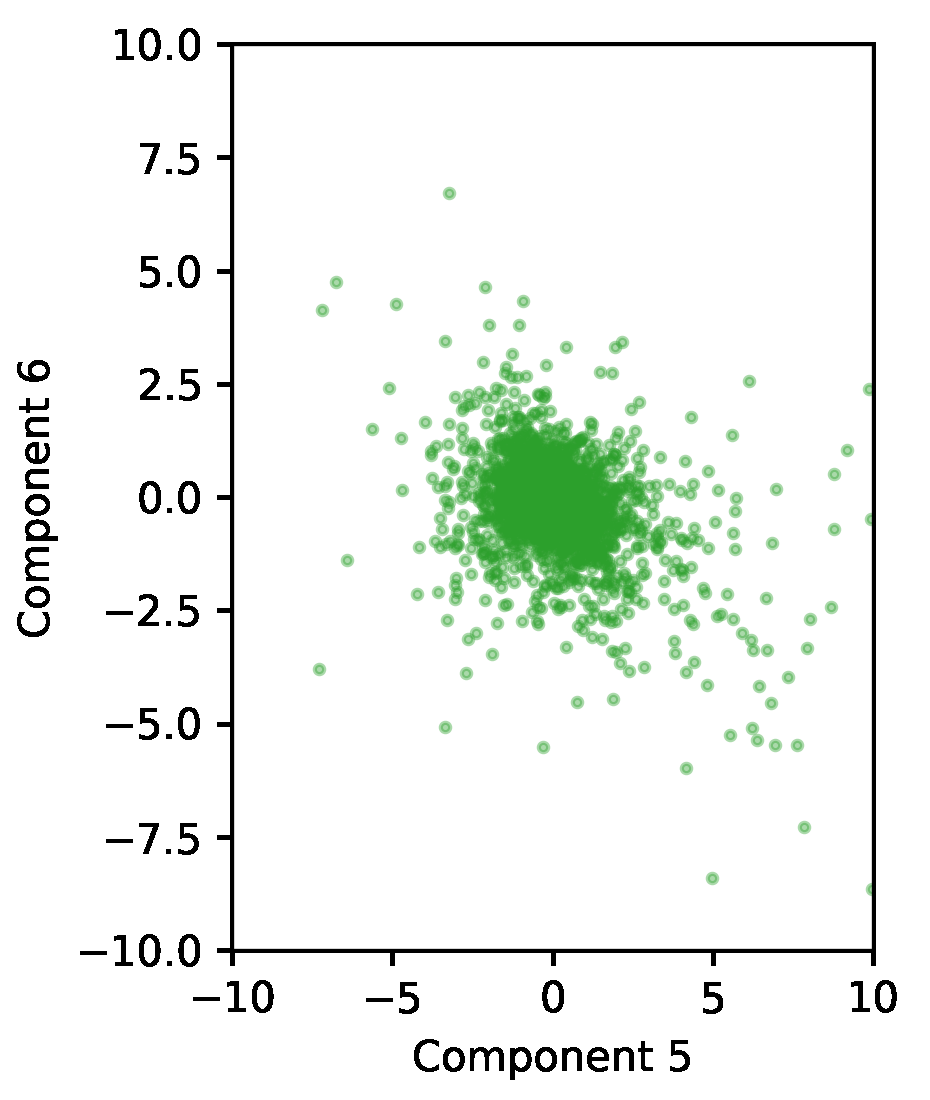}
    \end{subfigure}
    \hfill

    \begin{subfigure}{0.201\textwidth}
        \includegraphics[width=\textwidth]{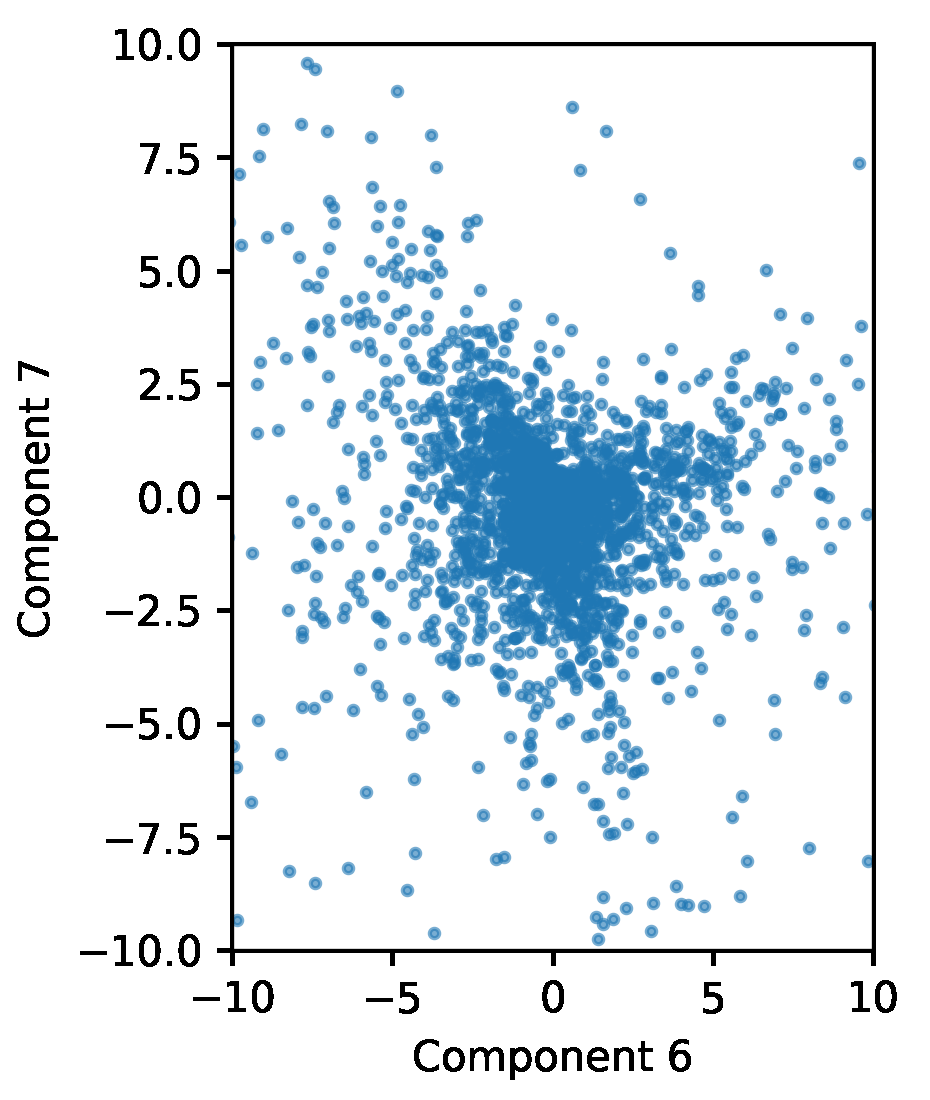}
    \end{subfigure}
    \begin{subfigure}{0.201\textwidth}
        \includegraphics[width=\textwidth]{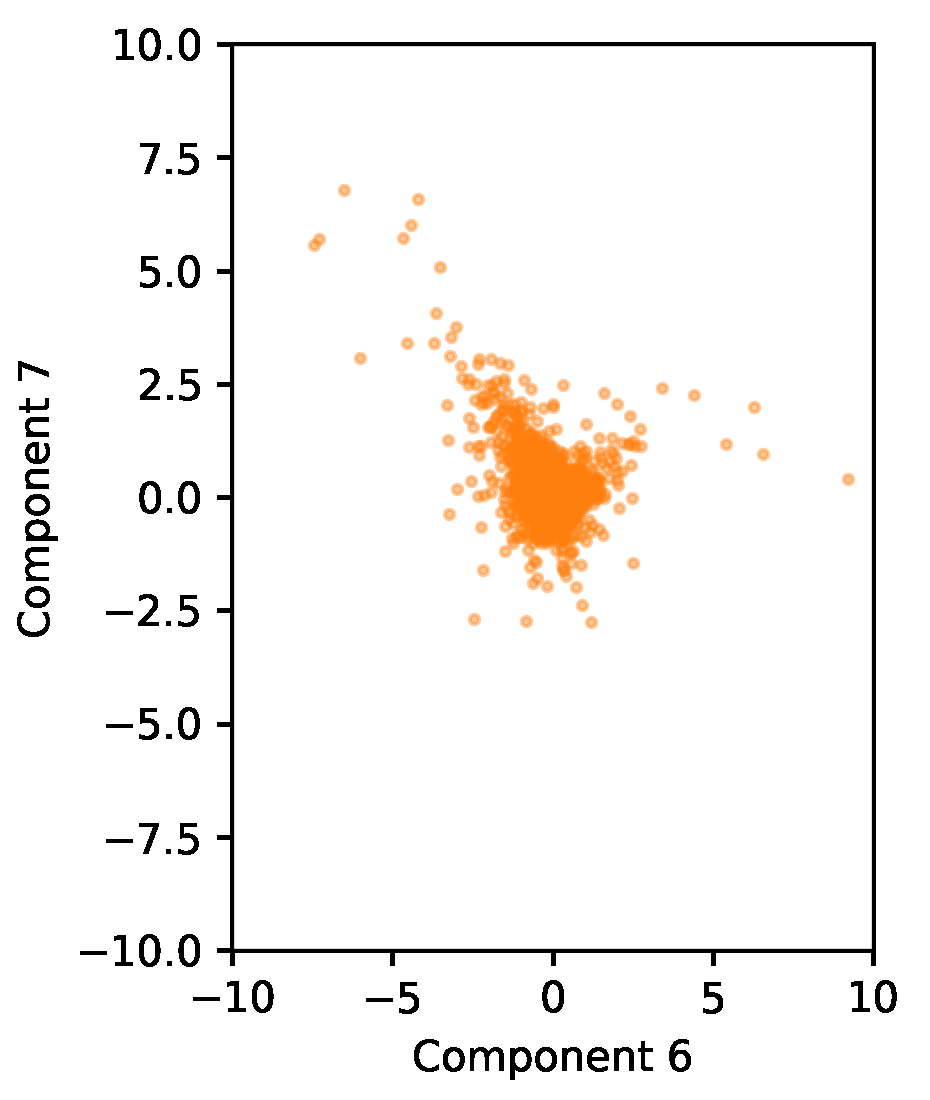}
    \end{subfigure}
    \begin{subfigure}{0.201\textwidth}
        \includegraphics[width=\textwidth]{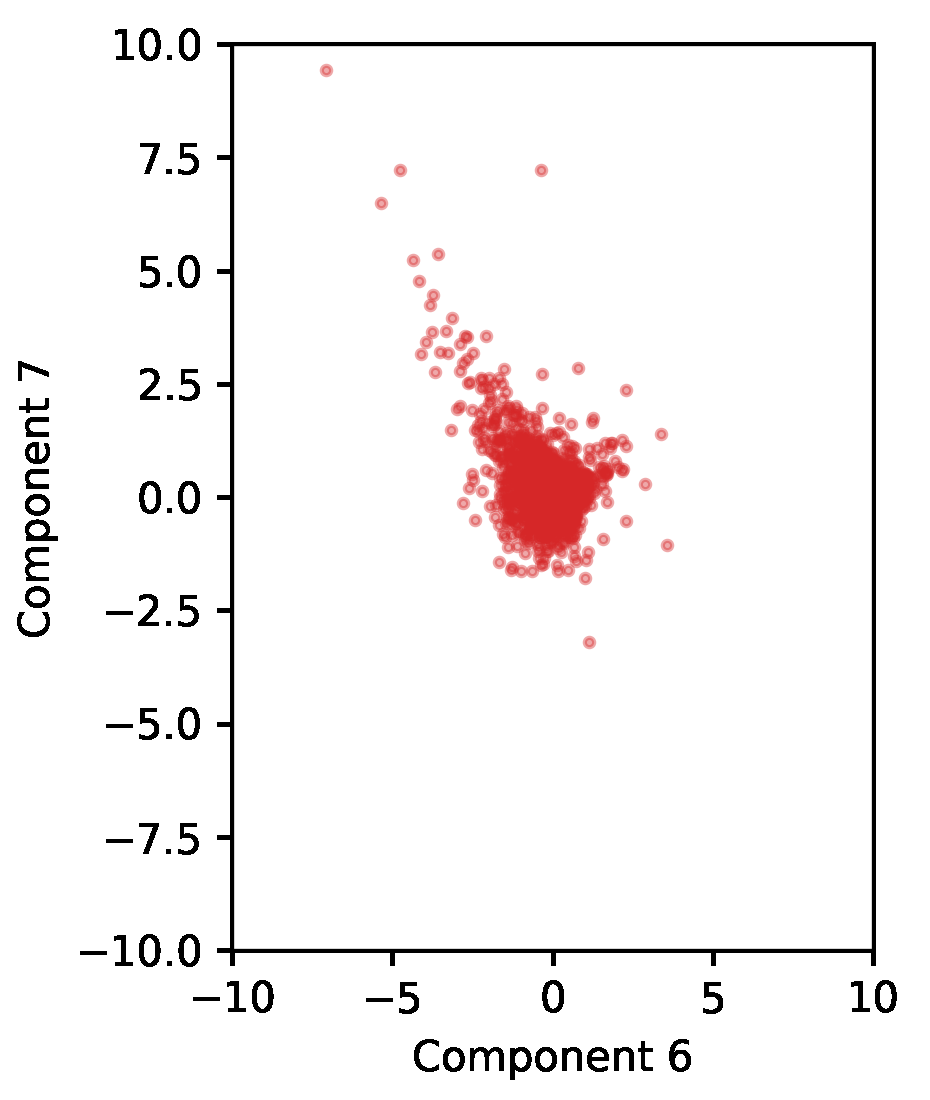}
    \end{subfigure}
    \begin{subfigure}{0.201\textwidth}
        \includegraphics[width=\textwidth]{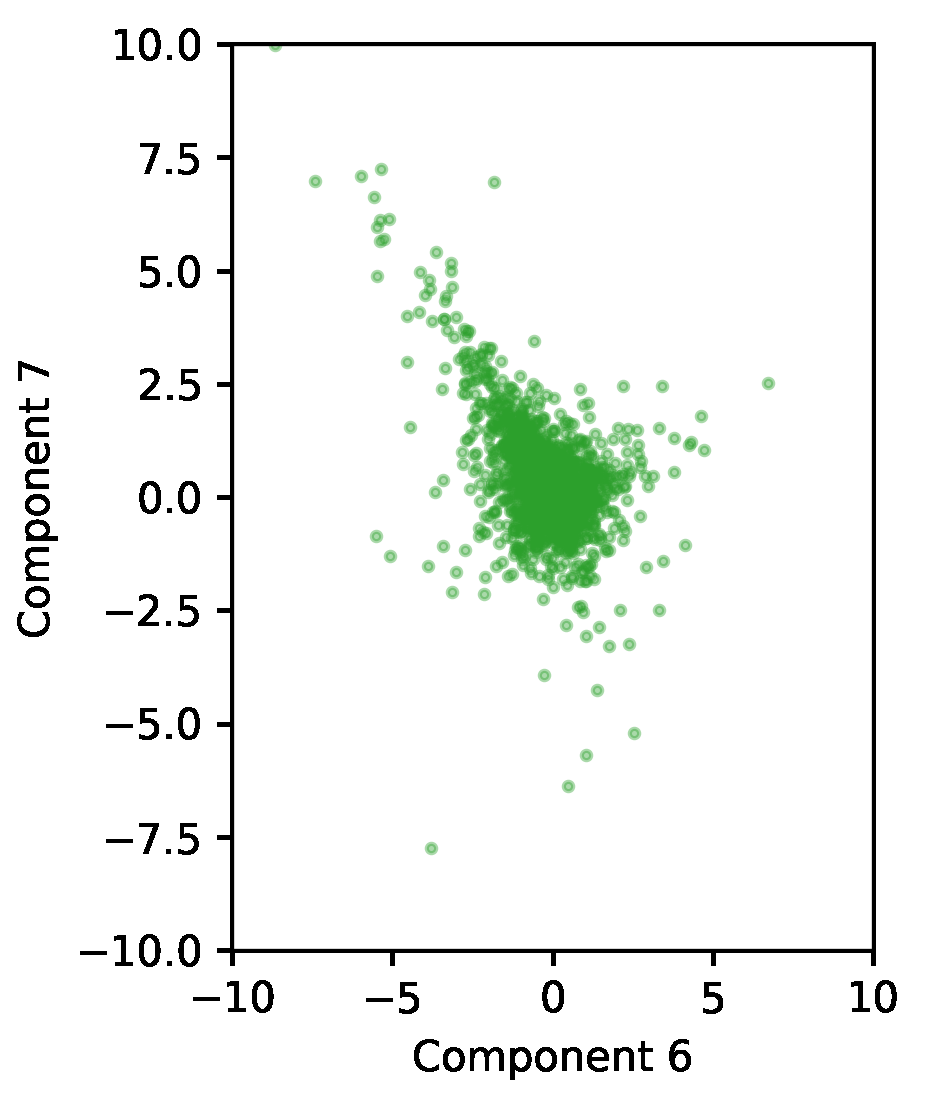}
    \end{subfigure}
    \hfill

    \begin{subfigure}{0.201\textwidth}
        \includegraphics[width=\textwidth]{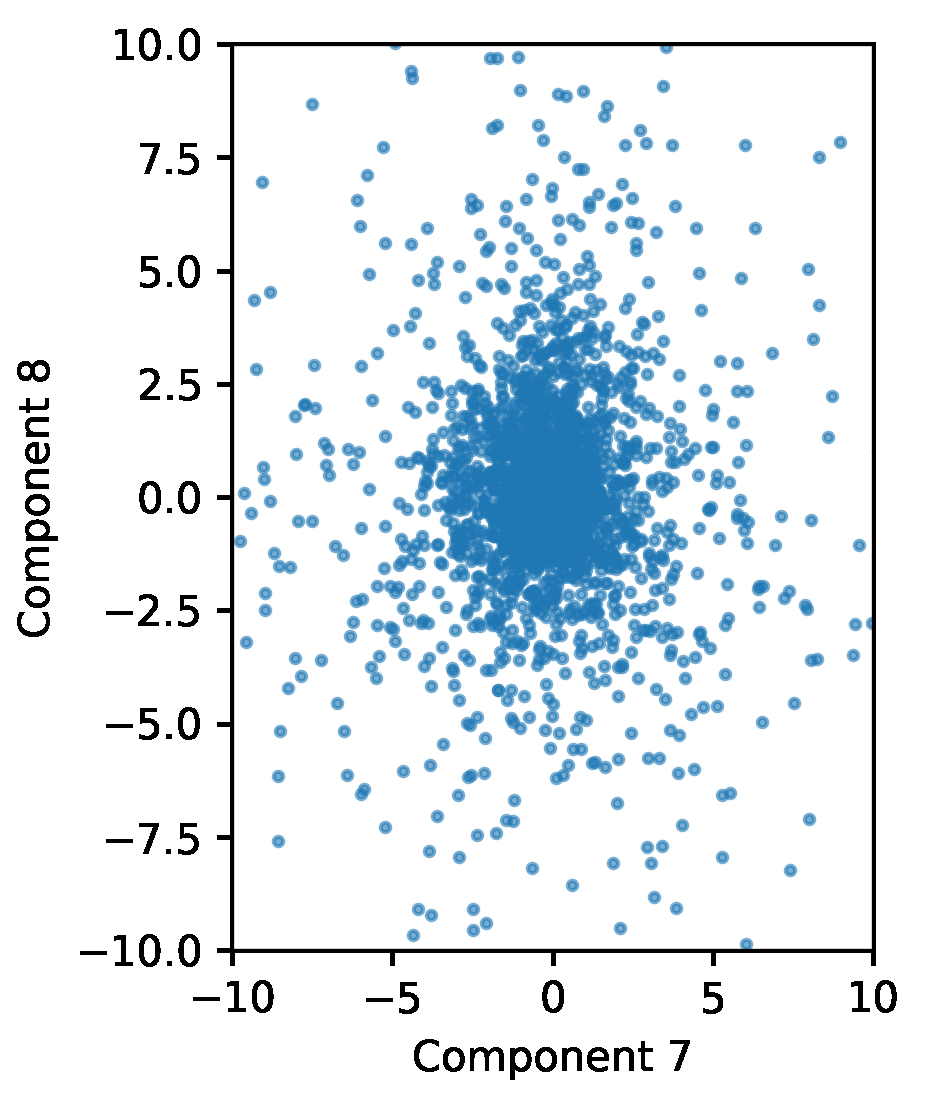}
    \end{subfigure}
    \begin{subfigure}{0.201\textwidth}
        \includegraphics[width=\textwidth]{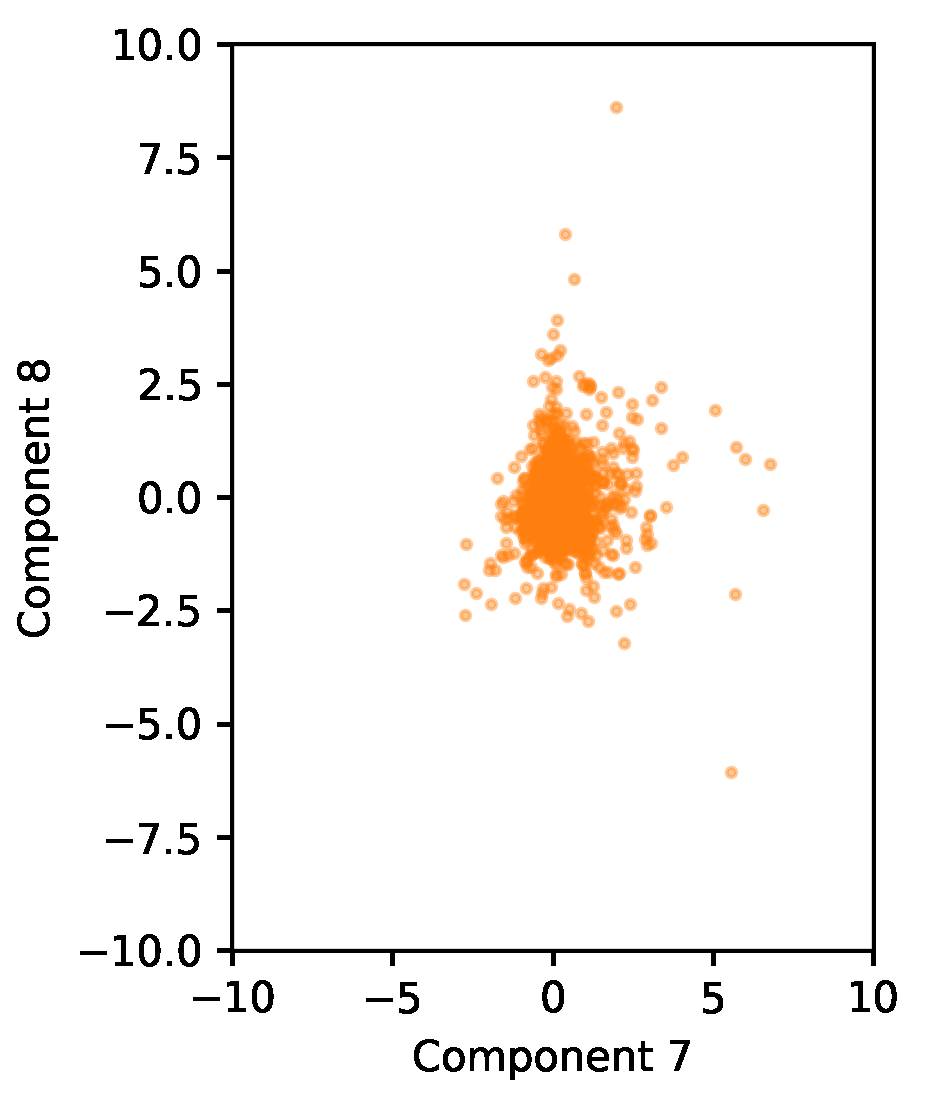}
    \end{subfigure}
    \begin{subfigure}{0.201\textwidth}
        \includegraphics[width=\textwidth]{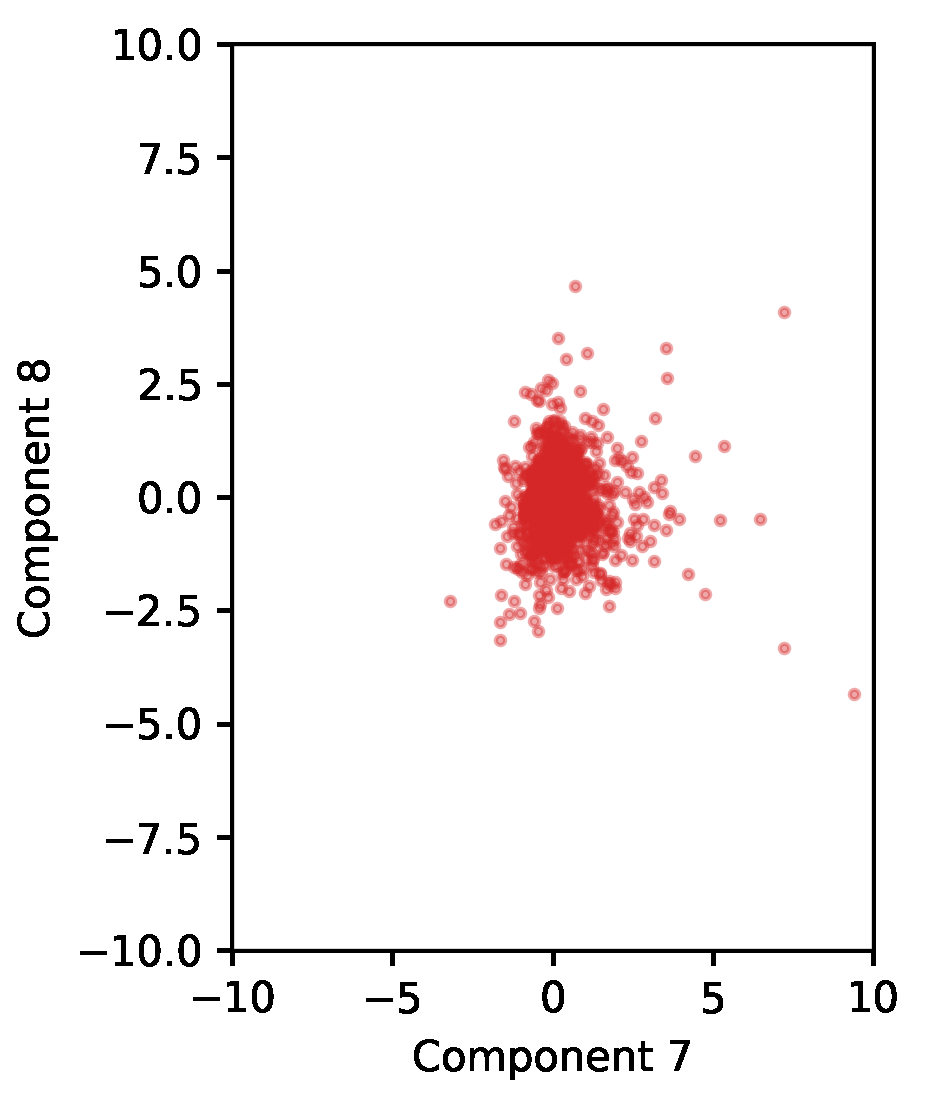}
    \end{subfigure}
    \begin{subfigure}{0.201\textwidth}
        \includegraphics[width=\textwidth]{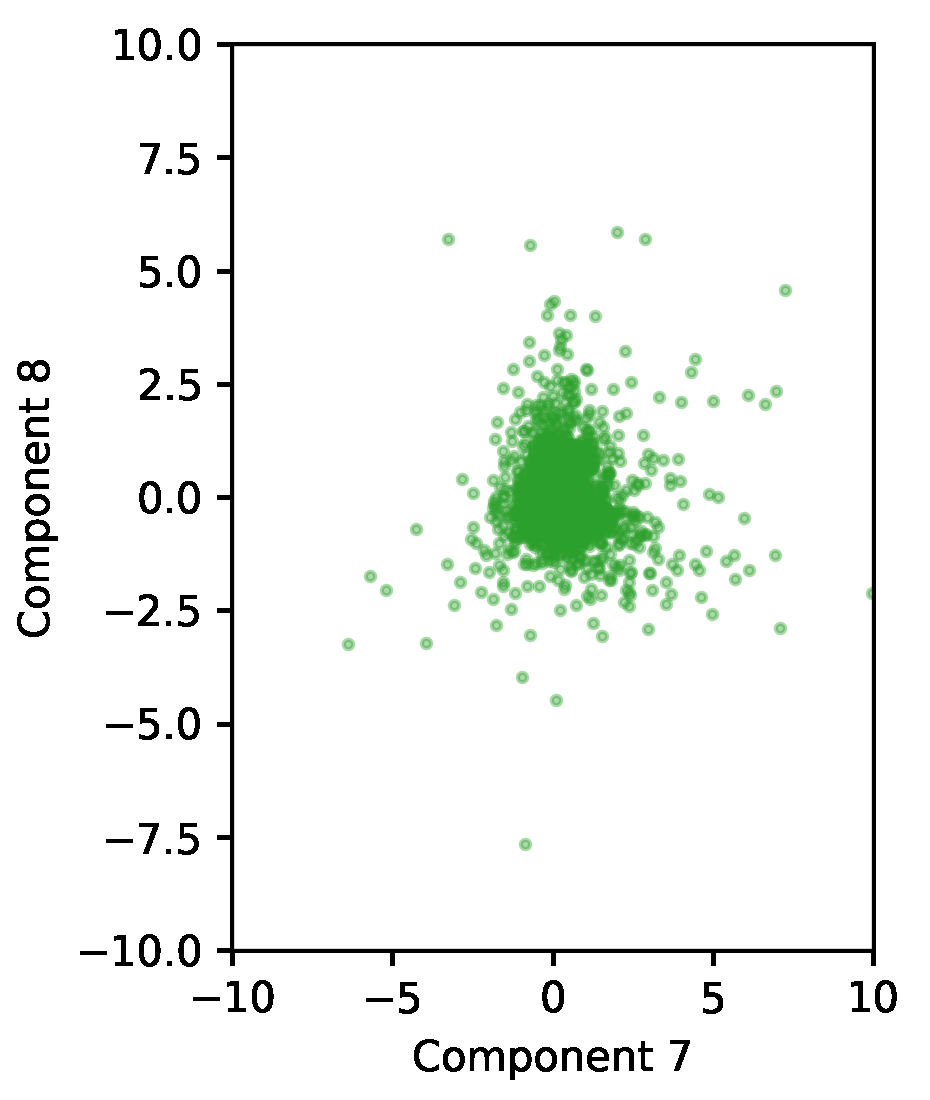}
    \end{subfigure}
    \hfill
    
    \begin{subfigure}{0.201\textwidth}
        \includegraphics[width=\textwidth]{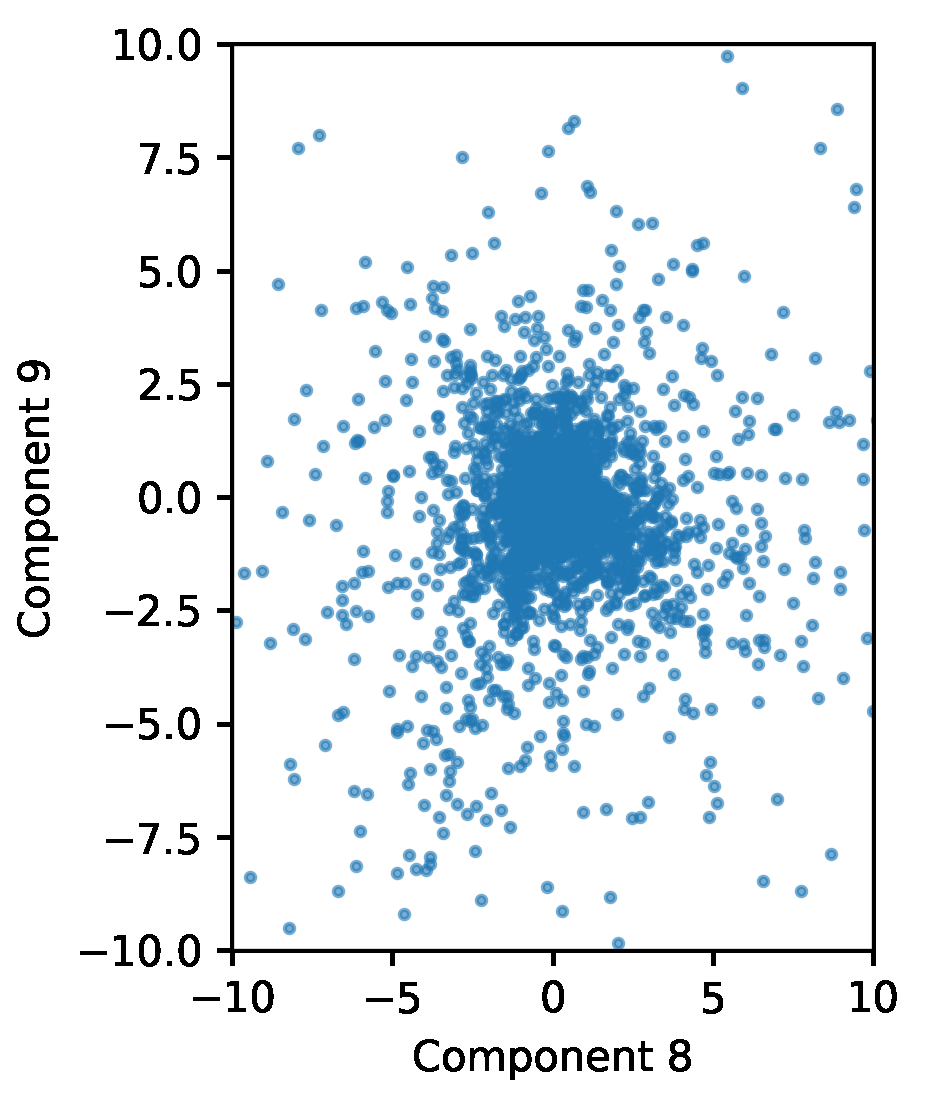}
        \caption{Webvicob}
    \end{subfigure}
    \begin{subfigure}{0.201\textwidth}
        \includegraphics[width=\textwidth]{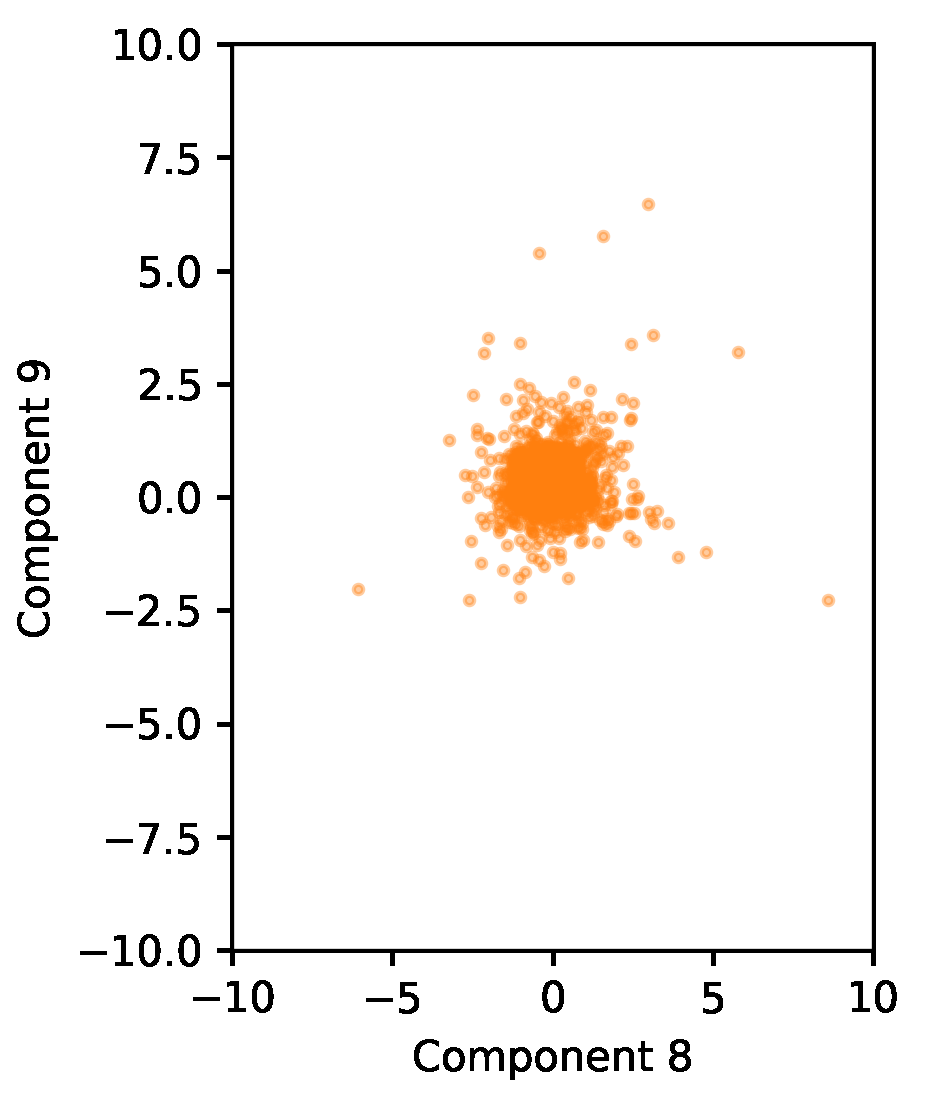}
        \caption{IIT-CDIP}
    \end{subfigure}
    \begin{subfigure}{0.201\textwidth}
        \includegraphics[width=\textwidth]{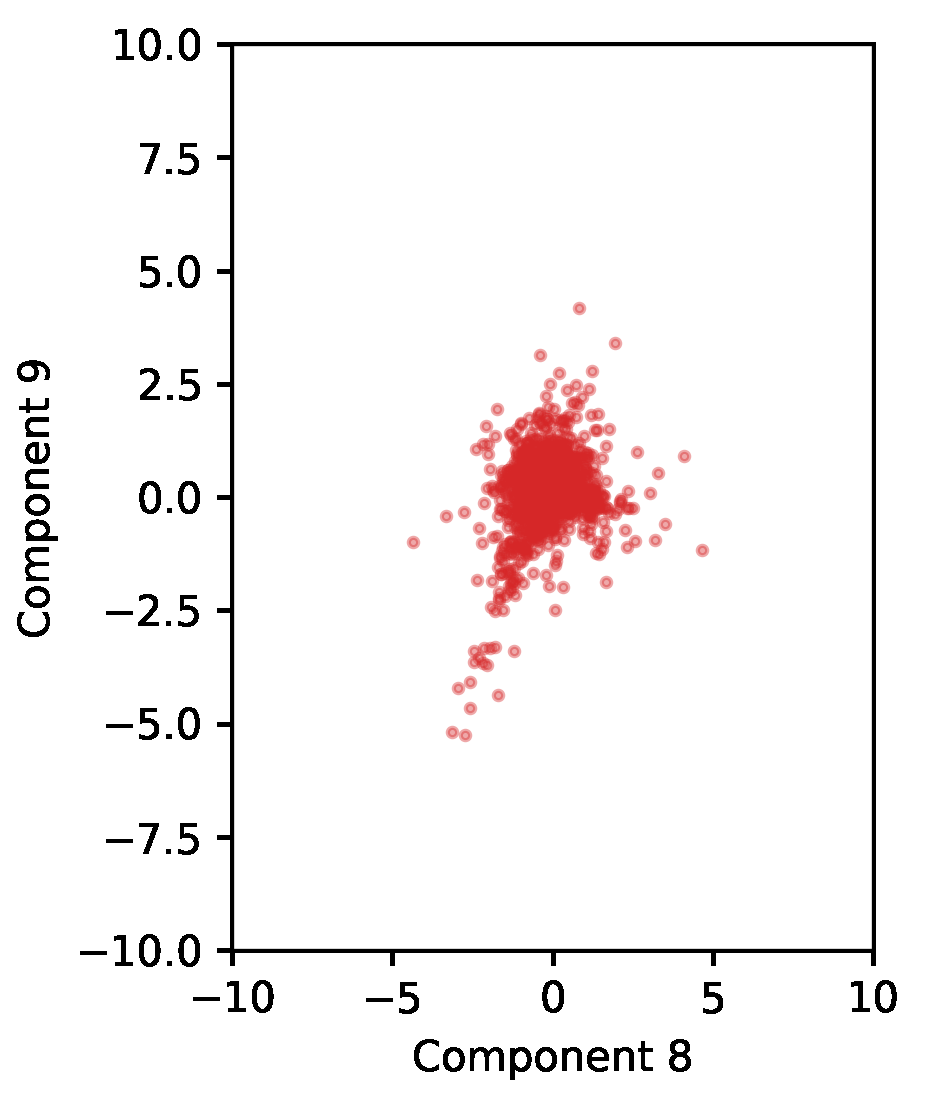}
        \caption{DocVQA Task 1}
    \end{subfigure}
    \begin{subfigure}{0.201\textwidth}
        \includegraphics[width=\textwidth]{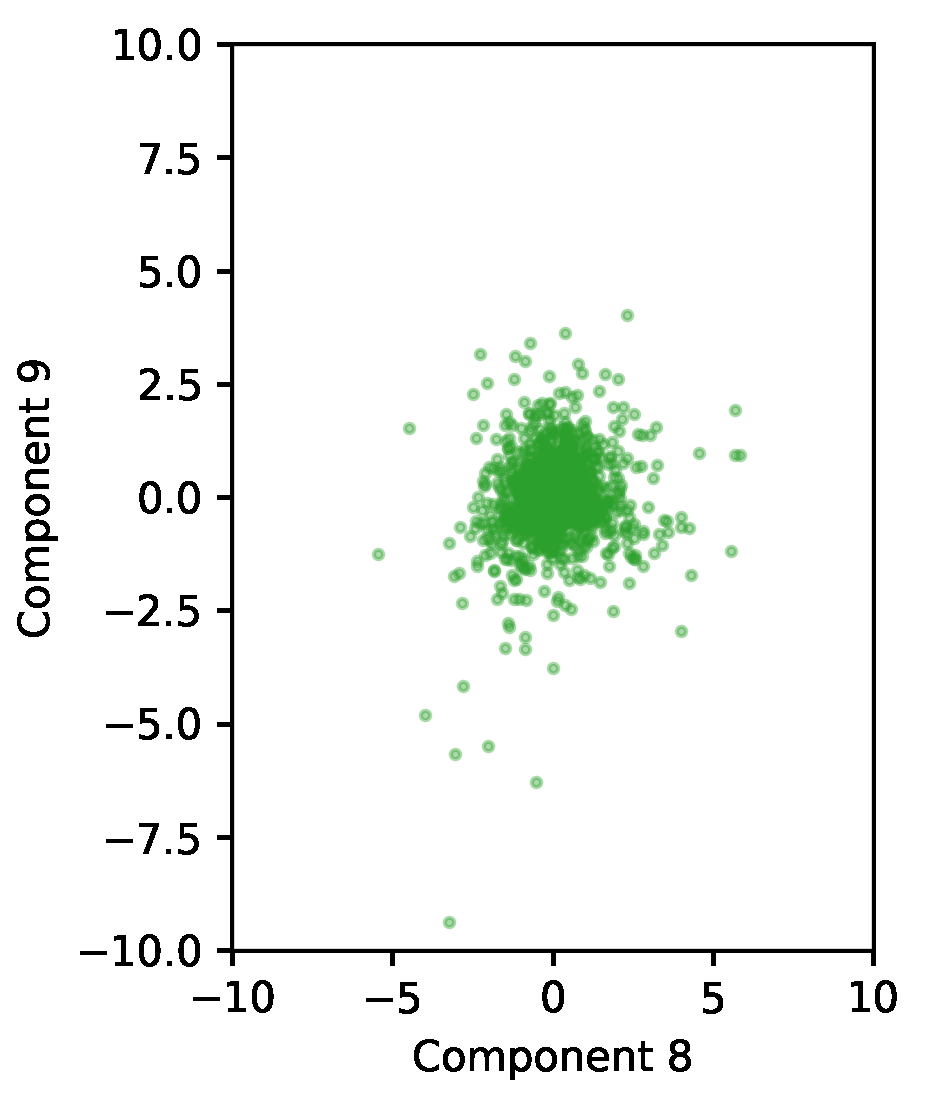}
        \caption{DocVQA Task 3}
    \end{subfigure}
    \hfill
    
    \caption{\textbf{Visualization of PCA Results.} Components 4, 5, 6, 7, 8, 9}
    \label{fig:pca2}
\end{figure}

We sampled 3,626 data points from each of the four datasets (Webvicob, IIT-CDIP, DocVQA Task 1, and DocVQA Task 3), creating a total of 14,504 vectors for PCA analysis.
We constructed a vocabulary of 100,000 words using Wikipedia 1M data and represented the data using Bag of Words (BOW) representation, excluding stop words\footnote{\scriptsize \url{https://www.nltk.org/index.html}}.
The figures [\ref{fig:pca1}, \ref{fig:pca2}] in the $i^{th}$ row display the analysis between the $i^{th}$ and $(i+1)^{th}$ principal components. For example, the first row shows the plot of the $0^{th}$ and $1^{st}$ principal components. 

We visualized the data using 10 principal components and found that the BOW vectors of IIT-CDIP and DocVQA Task 1 are very similar, while Webvicob contains a diverse range of BOW vectors.

\begin{figure}[htp]
    \centering
    \includegraphics[width=\textwidth]{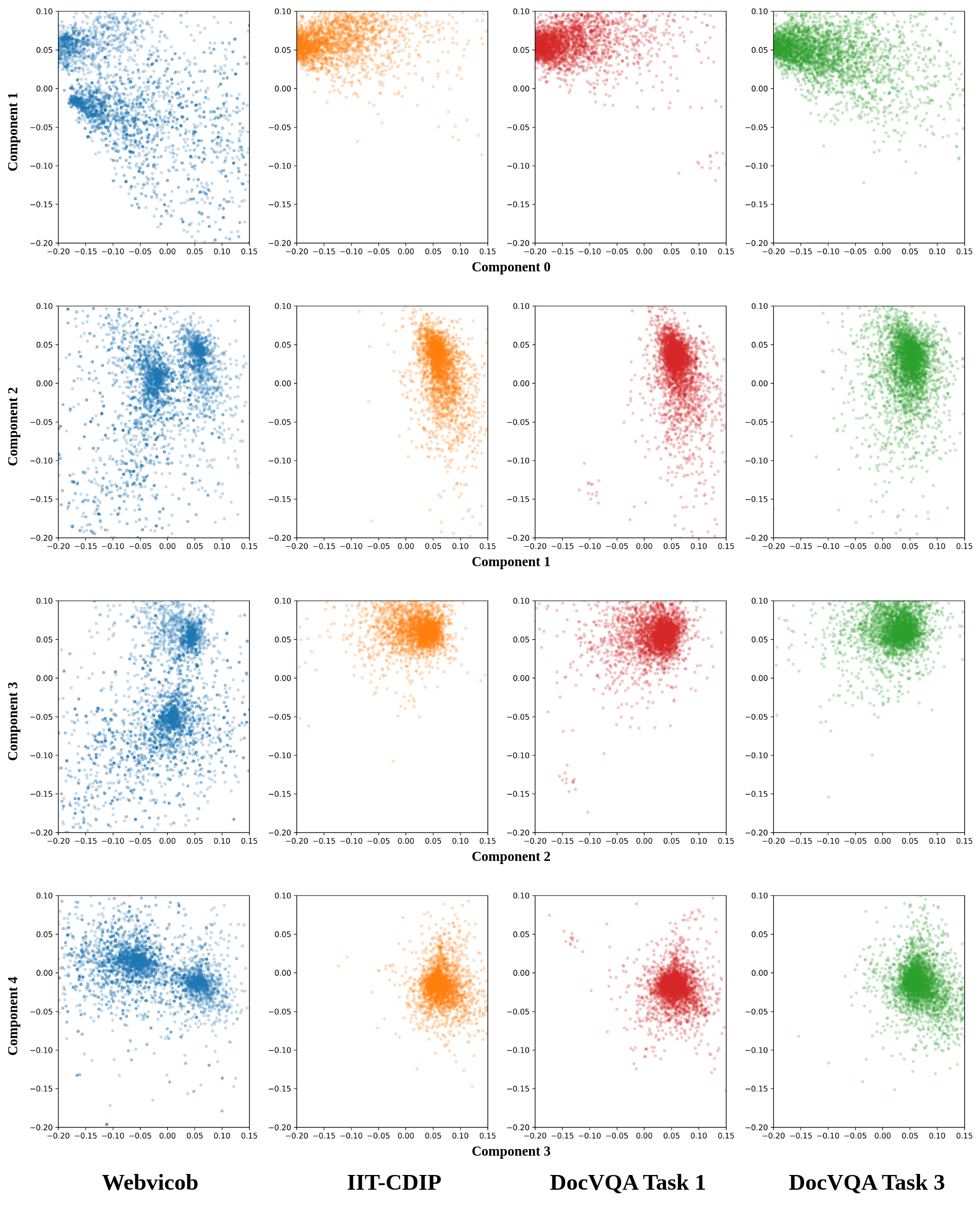}
    \caption{\textbf{Visualization of PCA Results with Google vocabulary.} Components 0, 1, 2, 3, 4}
    \label{fig:pca_google1}
\end{figure}

\begin{figure}[htp]
    \centering
    \includegraphics[width=\textwidth]{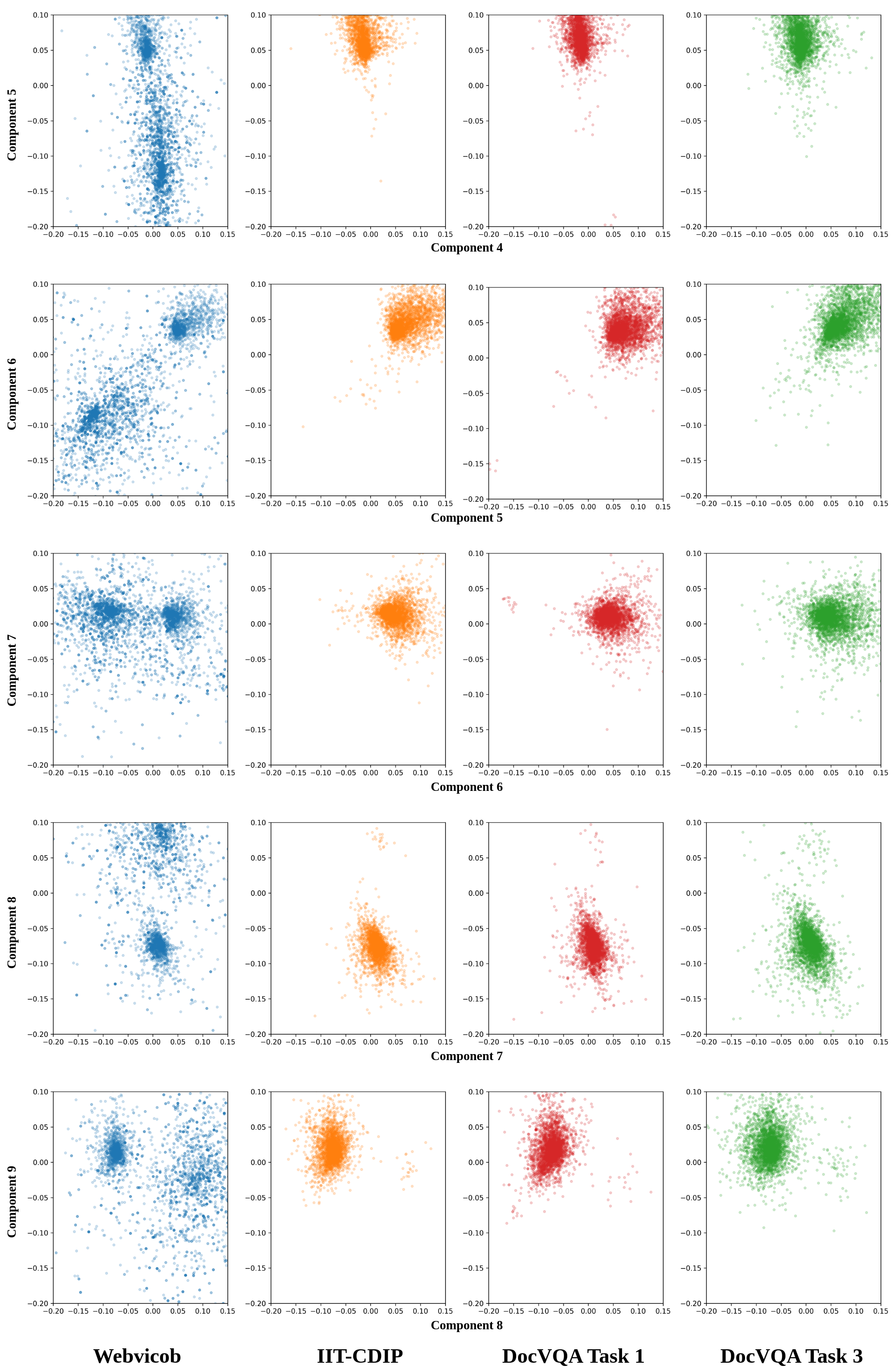}
    \caption{\textbf{Visualization of PCA Results with Google vocabulary.} Components 4, 5, 6, 7, 8, 9}
    \label{fig:pca_google2}
\end{figure}

Further visualization with Google vocabulary\footnote{\scriptsize \url{https://github.com/first20hours/google-10000-english/blob/master/google-10000-english-no-swears.txt}} can be found in figures [\ref{fig:pca_google1}, \ref{fig:pca_google2}].

\section{Conclusion}
In this work, we propose an engine, Webvicob, that builds visual corpora from web resources.
The constructed visual corpora can be utilized in building VDU backbones.
In our experiments and analyses, we observe that the Webvicob-generated data helps the VDU backbone perform robustly across a variety of document formats and domains.

\section{Acknowledgements}
The authors thank NAVER CLOVA Text Vision Team and Information Extraction Team.

\bibliographystyle{splncs04}
\bibliography{custom}

\end{document}